\documentclass[aps,twocolumn, superscriptaddress,floatfix]{revtex4-2}
\usepackage{xcolor}
\usepackage{amsmath, amssymb, amsthm}
\usepackage{graphicx}
\usepackage{hyperref}
\usepackage{tikz}
\usetikzlibrary{quantikz}
\usepackage{subfig}
\usepackage{booktabs}
\usepackage{orcidlink}
\usepackage{makecell}

\hypersetup{
    colorlinks=true,
    linkcolor=blue,
    filecolor=magenta,
    urlcolor=cyan,
    citecolor=blue,
}

\begin{document}

\title{Universal Quantum Transformer \\for Exact Reasoning in Arithmetic, Algebra, and Linguistics}

\author{Sungyong Chung~\orcidlink{0009-0001-3131-6011}}
\affiliation{Grainger College of Engineering, Department of Civil and Environmental Engineering, University of Illinois Urbana-Champaign}

\author{Alireza Talebpour~\orcidlink{0000-0002-5412-5592}}
\email{ataleb@illinois.edu}
\affiliation{Grainger College of Engineering, Department of Civil and Environmental Engineering, University of Illinois Urbana-Champaign}

\begin{abstract}
\noindent \textcolor{black}{Classical continuous-space neural networks can exhibit unstable generalization on exact reasoning tasks, including modular arithmetic, non-Abelian group algebra, and systematic compositional generalization. On such tasks, delayed generalization dynamics known as grokking do not necessarily produce a persistent zero-variance regime at 100\% test accuracy.} Here, we introduce the Universal Quantum Transformer (UQT), a novel, quantum-native computing architecture that uses the physical properties of multi-qubit systems as a universal inductive bias for exact algebraic \textcolor{black}{and compositional} reasoning. Rather than translating classical neural mechanisms, our framework relies entirely on parameterized geometric phase embedding and $SU(2)$ wave-interference. We demonstrate that \textcolor{black}{an identical} quantum attention circuit, operating on a highly compact \textcolor{black}{5 or 6} qubit substrate \textcolor{black}{with only 551 to 1,650 trainable parameters}, exactly learns \textcolor{black}{three} highly distinct formal classes: cyclic modular arithmetic ($\mathbb{Z}_{11}$), non-Abelian algebra (the $S_4$ permutation group)\textcolor{black}{, and systematic linguistic compositionality (the SCAN language).} \textcolor{black}{While the classical multi-layer perceptron (MLP) and Transformer baselines evaluated in this paper either fail to reach exact generalization or exhibit stochastic instability at convergence,} the UQT achieves mathematically exact, deterministic generalization. We define this stricter regime as crystallization: a step beyond the well-known phenomenon of grokking. Finally, we deploy the UQT on noisy intermediate-scale quantum (NISQ) hardware, \textcolor{black}{correctly classifying 39 of 40 selected instances on IBM Quantum computers.} \textcolor{black}{Relative to these tested baselines, the results indicate that the UQT provides a useful structural inductive bias for exact formal reasoning.}
\end{abstract}

\maketitle


\section{Introduction}

While attention-based models have achieved remarkable results in natural language processing, their capacity for \textcolor{black}{reasoning across cyclic arithmetic, non-Abelian group algebra, and systematic compositionality} remains brittle, with models often failing to generalize on structurally identical problems outside their training distribution \cite{dziri2023faith, trask2018neural, lake2018generalization}. Standard architectures operating in unconstrained continuous Euclidean space ($\mathbb{R}^n$) carry inductive biases misaligned with periodic, non-commutative, and compositional structure, not because such representations are theoretically impossible, but because they are not designed to handle such problems. As recent studies on algorithmic grokking reveal, utilizing continuous Euclidean space causes networks to default to pure memorization, struggling to generalize to unseen operands without extensive, delayed optimization \cite{power2022grokking, liu2022towards}. More importantly, even when classical networks finally achieve this delayed generalization (i.e., grokking), \textcolor{black}{they do not necessarily enter a persistent zero-variance regime at 100\% test accuracy}. Fundamentally, theoretical analyses \textcolor{black}{suggest} that the discrete, periodic structure of formal systems such as modular arithmetic is most naturally encoded through wave interference in phase space, a mechanism classical continuous-space models \textcolor{black}{generally} approximate through massive over-parameterization \cite{gromov2023grokking}. As \textcolor{black}{our experiments show in the following sections}, this approximation leaves classical Transformers vulnerable to persistent stochastic instability at convergence, i.e., even when they eventually grok these tasks, they still can oscillate \textcolor{black}{throughout training} without reaching deterministic stability. Such oscillations can be observed at standard classical multi-layer perceptrons (MLPs) with ${\sim}$1,000 parameters as well as Transformer structures with over 400,000 parameters. This suggests the deficiency is structural, not merely parametric.

The deficiency identified above is geometric in nature, as these formal systems possess intrinsic structural regularities that quantum systems natively embody, including the periodicity of cyclic arithmetic, the non-commutativity of group algebra, and the sequential compositionality of language. Specifically, quantum phase angles are $2\pi$-periodic, and $SU(2)$ operator composition is non-commutative, suggesting that a quantum system designed to exploit these structural properties could achieve exact formal reasoning where classical continuous-space models struggle. \textcolor{black}{The quantum machine learning literature has broadly pursued two strategies \cite{biamonte2017quantum, cerezo2022challenges, peral2024systematic}: translating classical architectures, such as neural networks, kernel methods, and attention mechanisms, into parameterized quantum circuits \cite{lakhdar2025benchmarking, zhang2025survey, li2024quantum}, or accelerating classical linear algebra through fault-tolerant quantum routines \cite{guo2024quantum, khatri2024quixer}.} \textcolor{black}{Because these conventional strategies treat quantum rotations as universal function approximators rather than as geometric structures deliberately aligned with the symmetries of the target formal system, they do not leverage quantum phase periodicity or operator non-commutativity as task-specific structural inductive biases. While recent equivariant quantum machine learning approaches have successfully aligned circuit structures with the symmetries of physical systems, such as molecular Hamiltonians or geometric point clouds \cite{larocca2022group, meyer2023exploiting}, extending these structural advantages to abstract sequential reasoning over discrete mathematical and linguistic spaces remains unexplored.}

The Universal Quantum Transformer (UQT) addresses this gap by mapping abstract symbols directly onto the unitary operations of a parameterized quantum system, leveraging these native quantum geometric properties as a universal physical inductive bias for exact mathematical and compositional reasoning. In this study, we establish that a single, unified topological modality, i.e., the quantum attention circuit, jointly embeds tokens into a shared superposition to natively resolve exact arithmetic\textcolor{black}{, non-Abelian group algebra, and systematic linguistic compositionality}.

\section{Universal Quantum Transformer}
\subsection{From grokking to crystallization}

To contextualize the physical advantage of the UQT, it is necessary to formally distinguish between the transient learning dynamics of algorithmic generalization and the structural stability of a model's converged representation. In the machine learning literature, the delayed discovery of generalizing solutions is broadly referred to as grokking \cite{power2022grokking, liu2022towards}.

\vspace{0.2cm}
\noindent \textbf{Definition 1: Grokking.} \textit{Let $t$ denote the optimization epoch, and let
$\mathcal{A}_{\mathrm{train}}(t)$ and $\mathcal{A}_{\mathrm{test}}(t)$ denote the training and test accuracies at epoch $t$, respectively. Let $t_{\mathrm{mem}}$ be the epoch at which the model perfectly memorizes the training dataset, i.e.,
$\mathcal{A}_{\mathrm{train}}(t_{\mathrm{mem}})=100\%$. Grokking is the phenomenon in which
$\mathcal{A}_{\mathrm{test}}(t)$ remains near random chance until a much later epoch
$t_{\mathrm{grok}} \gg t_{\mathrm{mem}}$, after which it rapidly increases toward $100\%$.}
\vspace{0.2cm}

However, because prior work has often treated grokking as a single, monolithic transition in generalization, it does not distinguish between approximate statistical generalization and exact recovery of the underlying mathematical rule. Achieving grokking merely guarantees that a neural network has found a statistical decision boundary that adequately covers the unseen data. When classical continuous-space Transformers learn algorithmic tasks, their generalization often emerges only after an extended memorization phase, and the resulting solution may remain an approximate statistical representation rather than an exact symbolic or algebraic resolution. Consequently, even when classical models exhibit grokking on the training distribution, their generalization may remain statistically fragile, as reflected by persistent accuracy oscillations and nonzero test-accuracy variance, i.e., $\operatorname{Var}[\mathcal{A}_{\mathrm{test}}(t)] > 0$, as we show in the following sections. To capture the ultimate resolution of formal mathematical systems, we introduce the stricter concept of crystallization.

\vspace{0.2cm}
\noindent \textbf{Definition 2: Crystallization.} \textit{A strict regime of generalization that requires, but supersedes, grokking. A network undergoes crystallization at epoch $t_c \ge t_{\mathrm{grok}}$ if $\mathcal{A}_{\mathrm{test}}(t_c) = 100\%$ and the model achieves deterministic stability, resulting in strictly zero variance for all subsequent epochs: $\operatorname{Var}[\mathcal{A}_{\mathrm{test}}(t)] = 0$ for all $t \ge t_c$.}

\vspace{0.2cm}

\textcolor{black}{In practice we verify this criterion over the training horizon reached in each experiment.} \textcolor{black}{Although classical models can in principle crystallize via over-parameterization \cite{gromov2023grokking}, the specific classical baselines evaluated here did not attain a persistent zero-variance state within their training horizons.}

\subsection{Quantum attention circuit}
\textcolor{black}{The UQT encodes \textcolor{black}{input token sequences (discrete symbols)} directly into a shared multi-qubit quantum state through sequential unitary composition, enabling the physical wave-interference of quantum mechanics to evaluate relational structure.} \textcolor{black}{While classical Transformer architectures \cite{vaswani2017attention} implement attention through explicit Query-Key-Value dot products to compute pairwise token relevance via learned projections, the UQT operates through a physically distinct mechanism.} Specifically, in the quantum attention circuit (Figure \ref{fig:circuit_parallel}), a sequence of $S$ input tokens ($x_1, \dots, x_S$) is mapped through an embedding table to a set of physical rotation angles that parameterize unitary transformations on an $N_{emb}$-qubit subspace of the total $N$-qubit register. These parameterized unitaries are applied consecutively (from $U(\vec{\theta}_{x_1})$ to $U(\vec{\theta}_{x_S})$) entirely prior to any explicit logical mixing operations. Due to the intrinsic compositional structure of quantum mechanics, where unitary operators combine via matrix multiplication, this sequence does not merely stack independent transformations but produces a joint operation that encodes the interaction between operands \textcolor{black}{implicitly} in the geometry of the resulting quantum state. The multi-qubit wave function therefore acts as a native geometric calculator \textcolor{black}{for operand composition}, in which the relative phases introduced by each operand interfere constructively or destructively. The resulting superposed state thus contains a distributed encoding of operand interactions, which is subsequently processed by $L$ deeper mixing layers to decode and extract task-relevant features.

As shown in Figure \ref{fig:circuit_micro}, a single mixing layer $U_{\mathrm{mix}}^{(l)}$ consists of parameterized single-qubit general $SU(2)$ Euler rotations ($U_{\mathrm{Rot}}^{(l)}$) applied to all $N$ qubits, followed by a global, cyclic conveyor-belt ring of CNOT gates. Because physical noisy intermediate-scale quantum (NISQ) hardware suffers severe decoherence when executing complex multi-qubit entangling operations, this hardware-efficient topology \textcolor{black}{progressively generates global entanglement} using only fundamental 2-qubit CNOT connections, providing a logical backbone capable of natively solving formal mathematical systems.

In this formulation, representation (phase-based embedding) is cleanly decoupled from reasoning (quantum interference followed by decoding), enabling a unified framework in which formal structure is evaluated through the native physics of the system. \textcolor{black}{This structural inductive bias is realized through gradient optimization, which organically aligns the circuit geometry with the target formal rules.}

Across our empirical evaluations, we instantiated the UQT on a highly compact \textcolor{black}{$N=5$ or $N=6$ qubit registers, depending on the experimental domain}. The total parameter capacity of the architecture is compact, defined strictly by the sum of the token embedding rotations and the \textcolor{black}{single-qubit $SU(2)$ rotation angles in the mixing layers.} For a token space of size $V$ embedded onto a subspace of $N_{emb}$ qubits (utilizing 4 rotations per qubit), and $L$ total mixing layers across all $N$ qubits, the parameter count is exactly $(V \times N_{emb} \times 4) + (L \times N \times 3)$. \textcolor{black}{In the configurations evaluated in this paper, the UQT uses 551--1,650 trainable parameters, compared with approximately 400,000 parameters in our Transformer baseline.} The complete mathematical formulation is provided in Appendix~\ref{app:math}.

\begin{figure*}[tb!]
    \centering

    \subfloat[Quantum attention circuit\label{fig:circuit_parallel}]{%
        \resizebox{0.75\linewidth}{!}{%
        \begin{quantikz}
        \lstick{$|q_0\rangle$} & \gate{U_{emb}(\vec{\theta}_{x_1,0})} & \qw \cdots & \gate{U_{emb}(\vec{\theta}_{x_S,0})} & \gate[6, style={fill=red!10}, nwires={2, 5}]{U_{\mathrm{mix}}^{(L)} \cdots U_{\mathrm{mix}}^{(1)}} & \meter{} \\
        \lstick{\vdots} & \vdots & \ddots & \vdots & & \vdots \\
        \lstick{$|q_{N_{emb}-1}\rangle$} & \gate{U_{emb}(\vec{\theta}_{x_1,N_{emb}-1})} & \qw \cdots & \gate{U_{emb}(\vec{\theta}_{x_S,N_{emb}-1})} & & \meter{} \\
        \lstick{$|q_{N_{emb}}\rangle$} & \qw & \qw \cdots & \qw & & \meter{} \\
        \lstick{\vdots} & \vdots & \ddots & \vdots & & \vdots \\
        \lstick{$|q_{N-1}\rangle$} & \qw & \qw \cdots & \qw & & \meter{}
        \end{quantikz}%
        }%
    }

    \vspace{0.6cm}

    \subfloat[Single mixing layer architecture ($U_{\mathrm{mix}}^{(l)}$)\label{fig:circuit_micro}]{%
        \resizebox{0.4\linewidth}{!}{%
        \begin{quantikz}
        \lstick{$|q_0\rangle$} & \gate{U_{\mathrm{Rot}}^{(l)}} & \ctrl{1} & \qw      & \qw      & \targ{}   & \qw \\
        \lstick{$|q_1\rangle$} & \gate{U_{\mathrm{Rot}}^{(l)}} & \targ{}  & \ctrl{1} & \qw      & \qw       & \qw \\
        \lstick{\vdots}        & \vdots                        & \qw      & \targ{}  & \ctrl{1} & \qw       & \qw \\
        \lstick{$|q_{N-1}\rangle$} & \gate{U_{\mathrm{Rot}}^{(l)}} & \qw      & \qw      & \targ{}  & \ctrl{-3} & \qw
        \end{quantikz}%
        }%
    }

    \caption{\textbf{Topology of the Universal Quantum Transformer (UQT).} The architecture strictly generalizes to arbitrary $N$-qubit registers and sequence lengths ($S$). \textbf{(a)} In the quantum attention circuit, $S$ sequential tokens ($x_1, \dots, x_S$) are embedded entirely prior to structural entanglement. The superposed phases are processed simultaneously by the $L$-layer mixing block. Measurements are performed on all $N$ qubits to generate the output. \textbf{(b)} A single layer $l$ of the shared mixing sequence, $U_{\mathrm{mix}}^{(l)}$, consisting of parameterized single-qubit general $SU(2)$ Euler gates ($U_{\mathrm{Rot}}^{(l)}$) and a global cyclic conveyor-belt ring of CNOT gates.}
    \label{fig:circuits}
\end{figure*}

\section{Results}
\subsection{Crystallization in modular arithmetic}
\begin{figure*}[tb!]
    \centering
    \subfloat[UQT: Mod 11 addition ($\mathbb{Z}_{11}$)\label{fig:mod11_addition_uqt}]{%
        \includegraphics[width=0.48\textwidth]{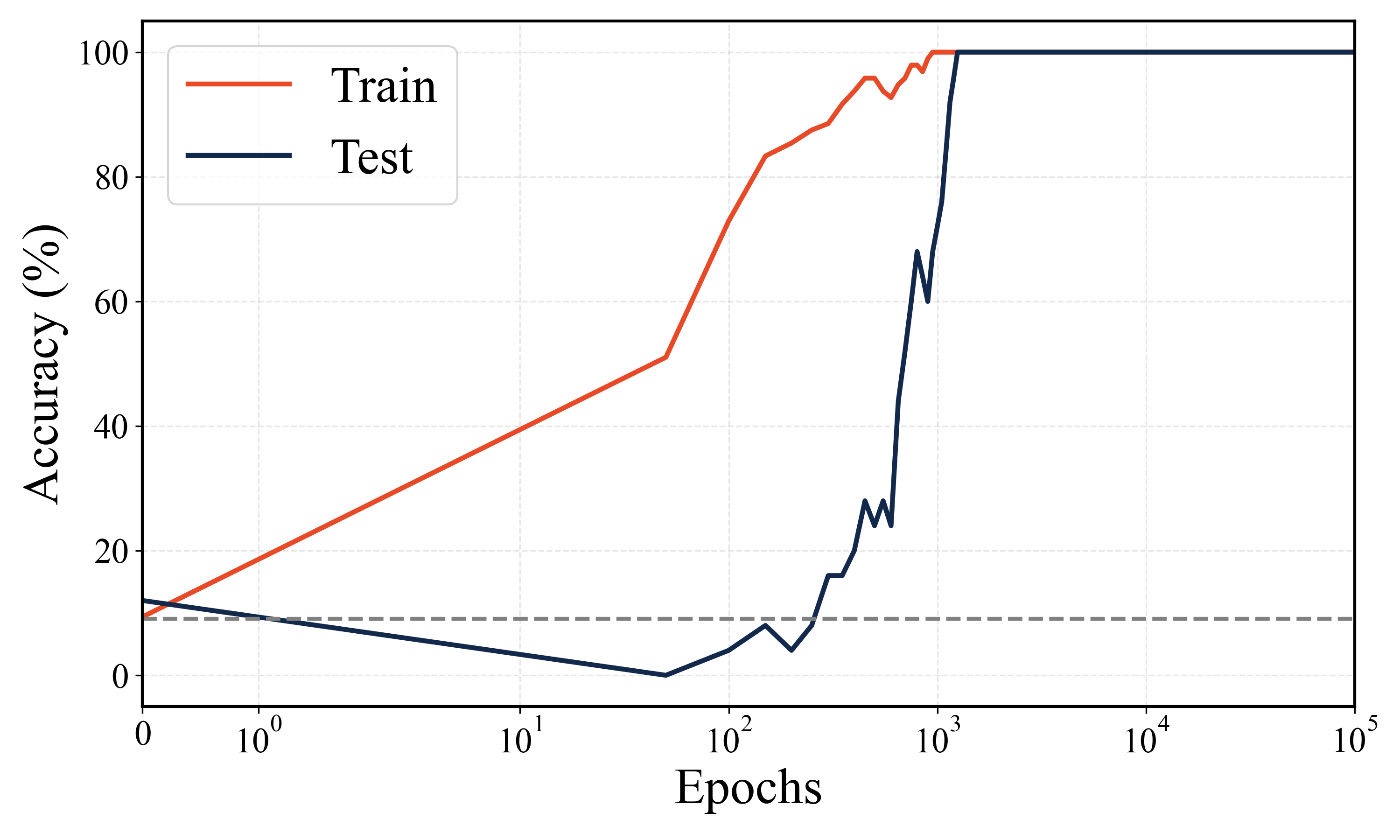}%
    }
    \hfill
    \subfloat[Transformer: Mod 11 addition ($\mathbb{Z}_{11}$)\label{fig:mod11_addition_classical_transformer}]{%
        \includegraphics[width=0.48\textwidth]{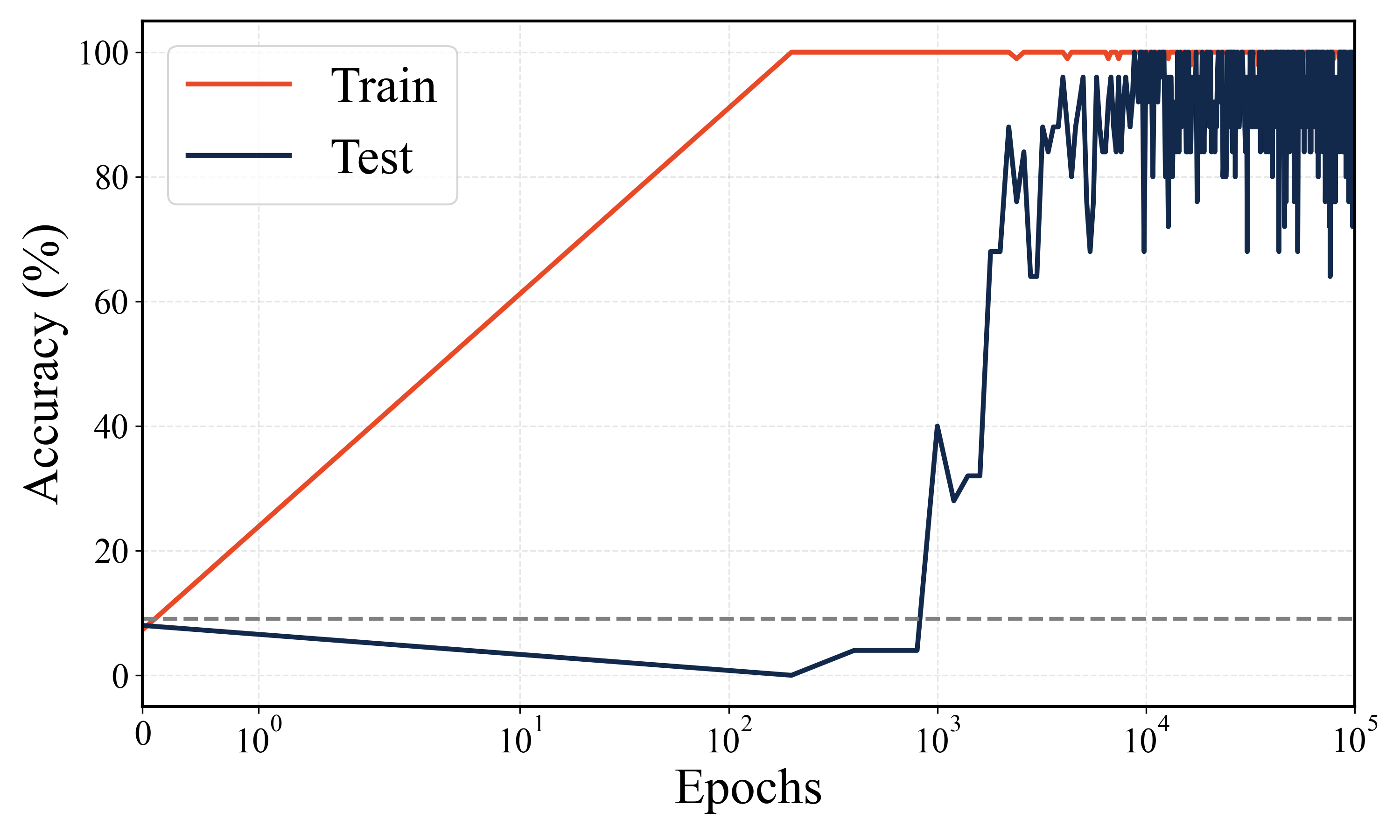}%
    }

    \vspace{0.4cm}

    \subfloat[UQT: Mod 11 multiplication ($\mathbb{Z}_{11}$)\label{fig:mod11_multiplication_with_zero_uqt}]{%
        \includegraphics[width=0.48\textwidth]{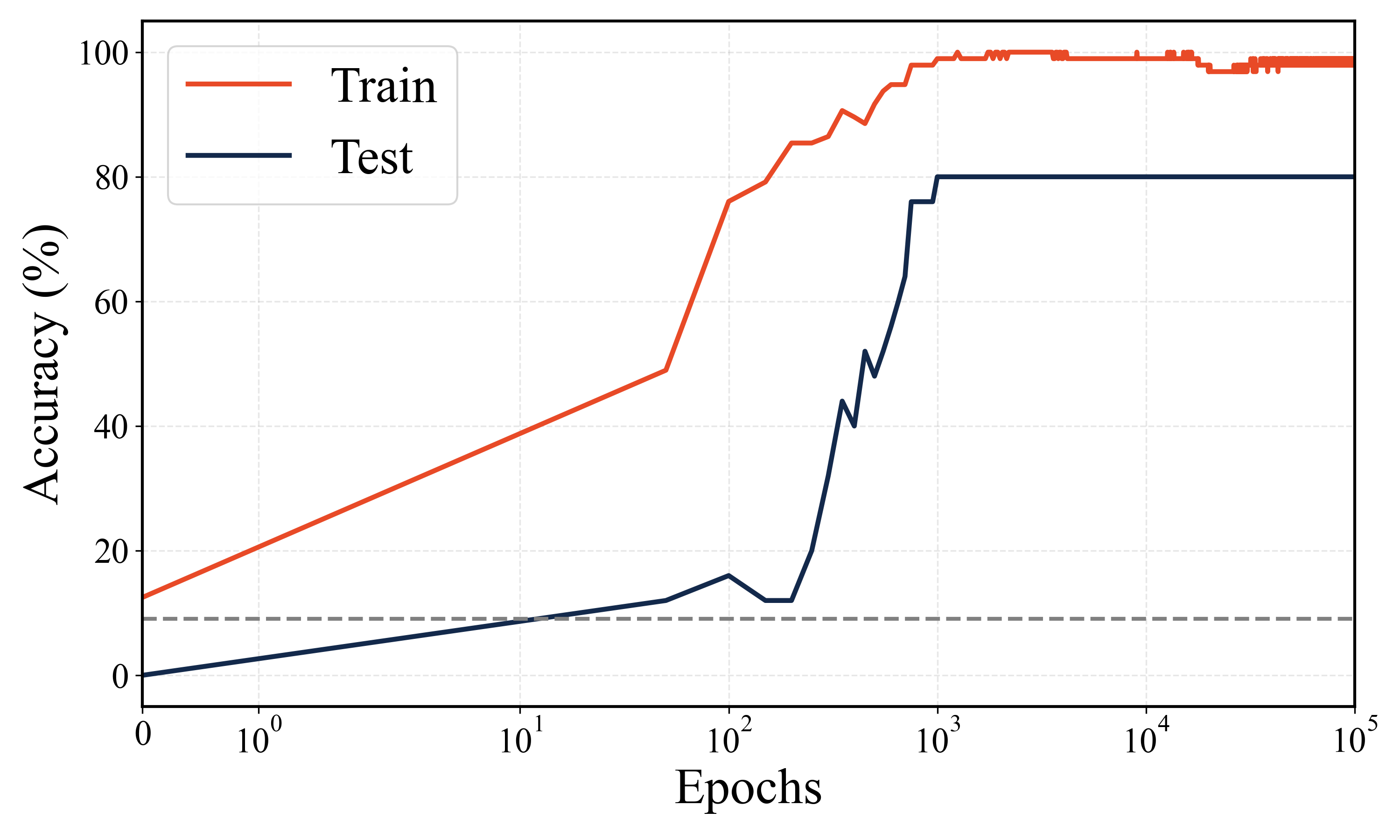}%
    }
    \hfill
    \subfloat[Transformer: Mod 11 multiplication ($\mathbb{Z}_{11}$)\label{fig:mod11_multiplication_with_zero_classical_transformer}]{%
        \includegraphics[width=0.48\textwidth]{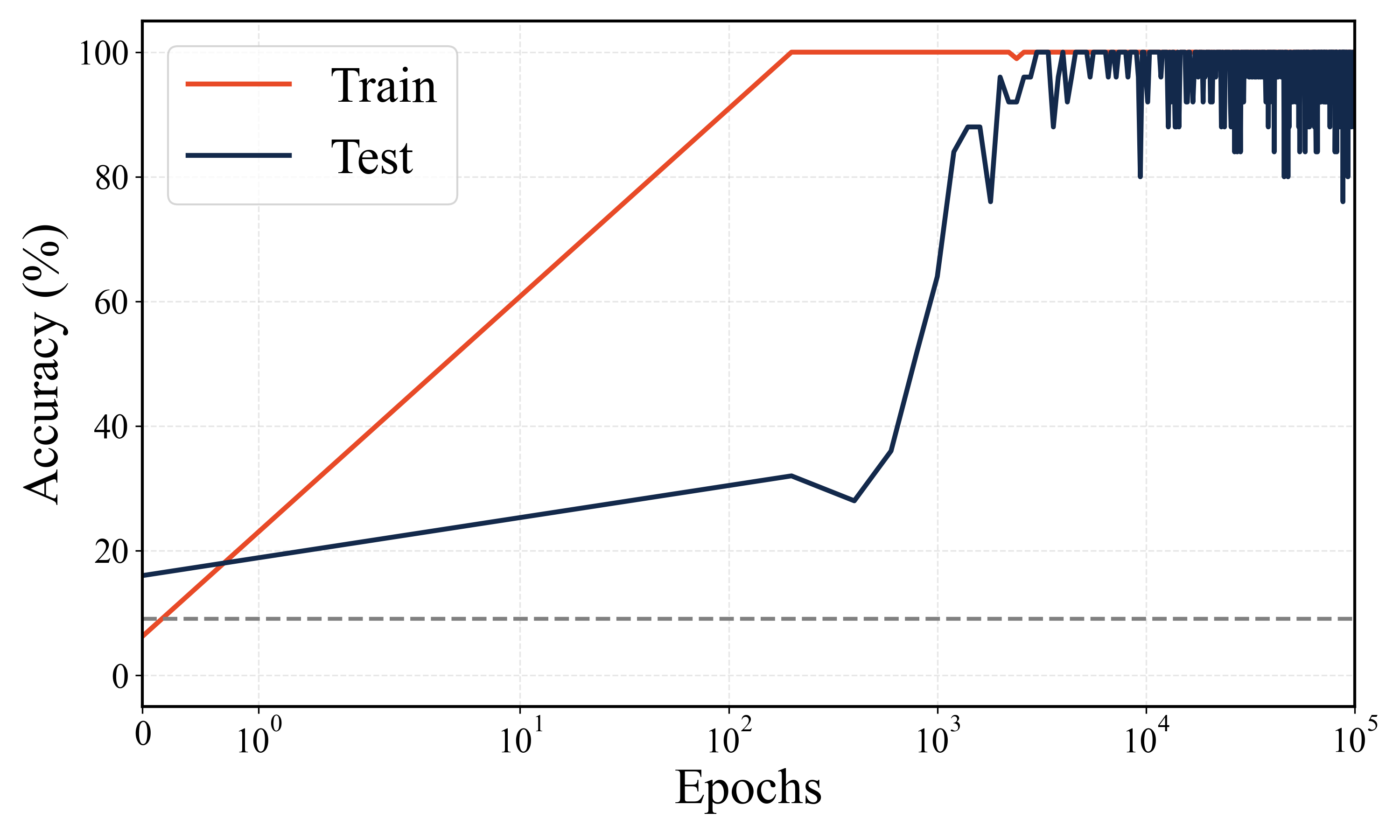}%
    }

    \vspace{0.4cm}

    \subfloat[UQT: Mod 11 multiplication ($\mathbb{Z}_{11}^*$)\label{fig:mod11_multiplication_no_zero_uqt}]{%
        \includegraphics[width=0.48\textwidth]{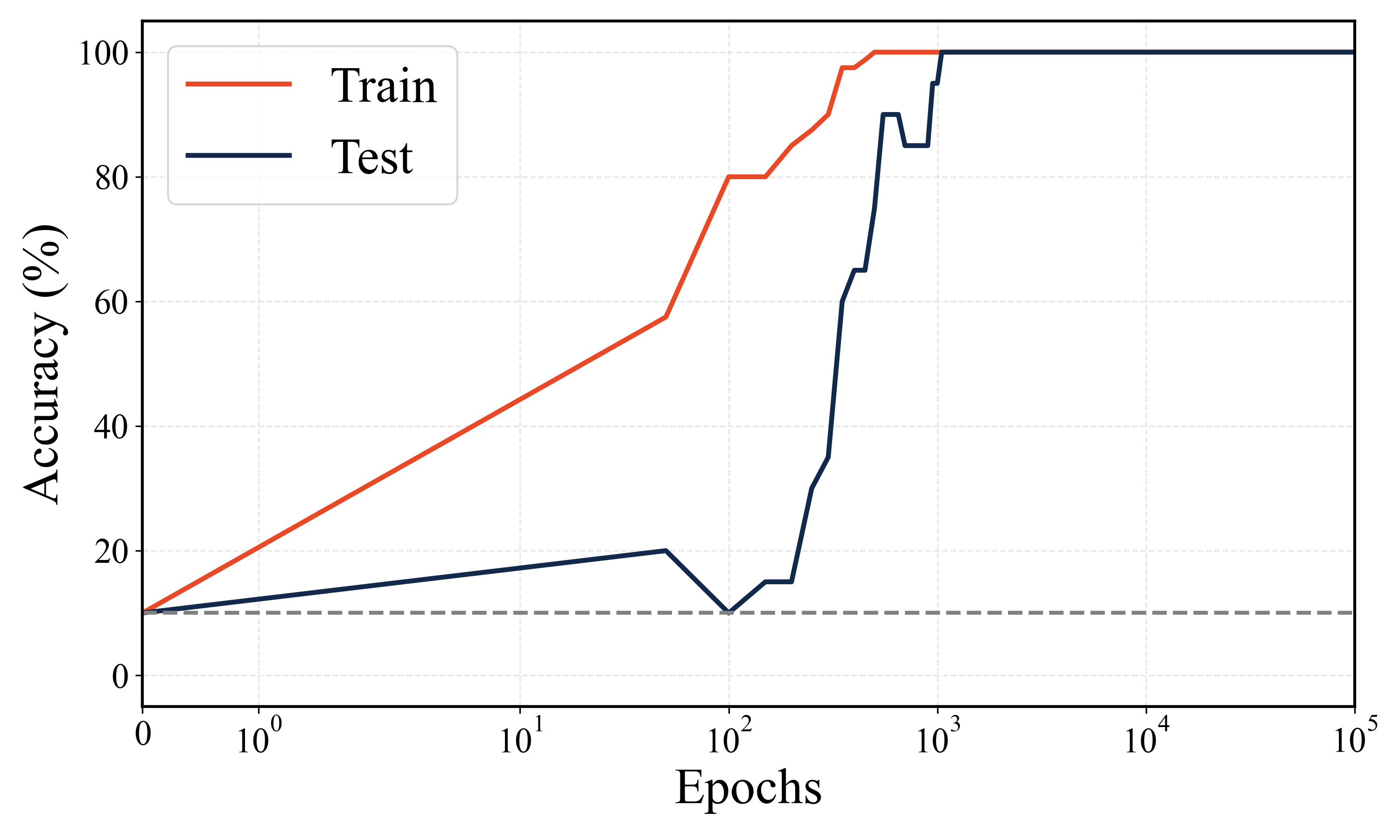}%
    }
    \hfill
    \subfloat[Transformer: Mod 11 multiplication ($\mathbb{Z}_{11}^*$)\label{fig:mod11_multiplication_no_zero_classical_transformer}]{%
        \includegraphics[width=0.48\textwidth]{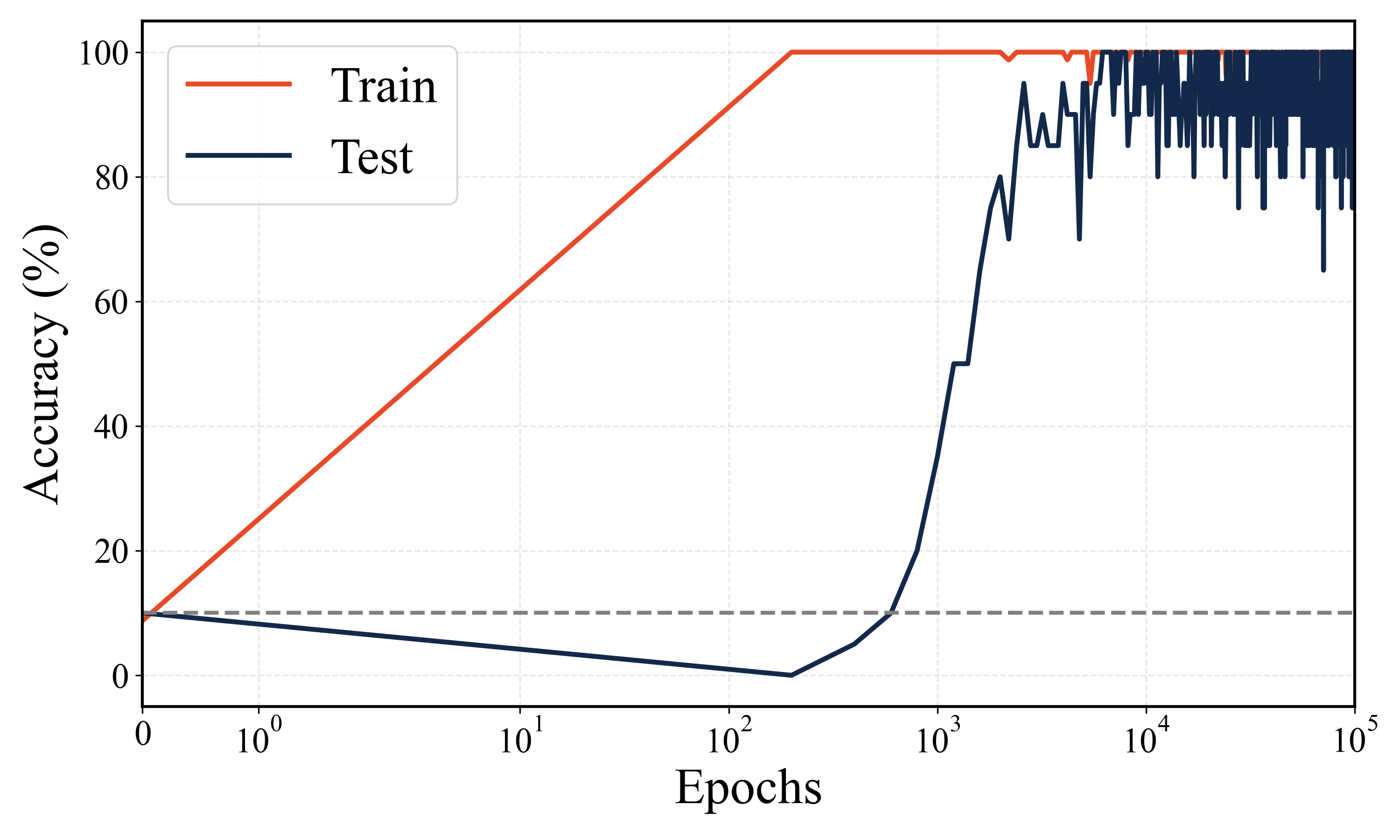}%
    }

    \caption{\textbf{Quantum crystallization in modular arithmetic.} \textbf{(a, b)} Both architectures learn addition, but the classical Transformer exhibits severe stochastic instability at convergence. \textbf{(c)} \textcolor{black}{The UQT fails to crystallize on multiplication with zero and does not learn the absorbing rule $x_1 \times 0 = 0$.} \textbf{(d)} The classical Transformer memorizes the zero-rule via pattern matching, buffering its instability. \textbf{(e, f)} When restricted to the $\mathbb{Z}_{11}^*$ multiplicative group (excluding the zero element), the UQT achieves exact, deterministic crystallization, whereas the classical network oscillates violently.}
    \label{fig:mod11_plots}
\end{figure*}

We tasked the quantum attention circuit with learning modular arithmetic given two input tokens, $x_1$ and $x_2$. To rigorously evaluate its geometric learning capabilities, we tested the architecture across three distinct algebraic regimes. First, we evaluated addition over the full prime Galois field $\mathbb{Z}_{11}$, where the network must predict $x_1 + x_2 \pmod{11}$ (Figures \ref{fig:mod11_addition_uqt} and \ref{fig:mod11_addition_classical_transformer}). Second, we evaluated multiplication over the full field $\mathbb{Z}_{11}$ ($x_1 \times x_2 \pmod{11}$), which includes the zero element (Figures \ref{fig:mod11_multiplication_with_zero_uqt} and \ref{fig:mod11_multiplication_with_zero_classical_transformer}). Finally, we evaluated multiplication restricted strictly to the bijective multiplicative group $\mathbb{Z}_{11}^* = \{1, 2, \dots, 10\}$, which excludes the zero element (Figures \ref{fig:mod11_multiplication_no_zero_uqt} and \ref{fig:mod11_multiplication_no_zero_classical_transformer}). In our implementation, we mapped the token space across an $N_{emb}=4$ qubit subspace and utilized a mixing depth of $L=25$. Measuring the full 5-qubit register ($N=5$), this architecture requires only 551 trainable parameters. Training hyperparameters and dataset construction are detailed in Appendix~\ref{app:impl}.

\textcolor{black}{Modular arithmetic can be challenging for generic neural architectures because its cyclic structure is not directly imposed by their parameterization, and prior work has documented delayed grokking on such tasks \cite{power2022grokking}.} To contextualize the performance of our architecture, we trained classical attention-based Transformers as a baseline (see Appendix~\ref{app:impl} for architecture details).

\textcolor{black}{The Transformer baseline shown in Figures \ref{fig:mod11_addition_classical_transformer}, \ref{fig:mod11_multiplication_with_zero_classical_transformer}, and \ref{fig:mod11_multiplication_no_zero_classical_transformer} contains approximately 400,000 trainable parameters.} Although \textcolor{black}{this baseline successfully groks each dataset, exhibiting a delayed transition in test accuracy long after reaching 100\% training accuracy, it does not maintain stable 100\% generalization through the end of training. Instead, its test accuracy exhibits severe fluctuations.} For instance, it easily isolates and memorizes the simple zero-rule ($x_1 \times 0 = 0 \pmod{11}$) during early epochs of training. This \textcolor{black}{behavior raises} the lower bound of the test accuracy, buffering the lowest dips of the oscillations to remain above 80\% (Figure \ref{fig:mod11_multiplication_with_zero_classical_transformer}). Once the zero element is excluded, the test accuracy oscillates between 70\% and 100\% (Figure \ref{fig:mod11_multiplication_no_zero_classical_transformer}).

In stark contrast to the classical model, the UQT cannot take the zero-rule shortcut. As demonstrated in Figure \ref{fig:mod11_multiplication_with_zero_uqt}, when the UQT attempts to learn the complete $\mathbb{Z}_{11}$ multiplication field (including the $0$ element), the network fails to reach 100\% accuracy. \textcolor{black}{One possible explanation is a structural mismatch between the absorbing element and the UQT encoding. The second operand acts as a unitary on the state already prepared by the first, so a faithful realization of $N \times 0 = 0$ would require $U_0 U_N = U_0$ for every $N$. Because $U_0$ is invertible, this forces every $U_N$ to be the identity, which cannot simultaneously represent the nonzero products, pointing to a limitation of the shared-register sequential unitary encoding used in the UQT.} Addition, however, \textcolor{black}{acts as an invertible translation for every operand, including zero ($x_1 + 0 = x_1$)}, allowing the UQT to crystallize (Figure \ref{fig:mod11_addition_uqt}).

When evaluated on the restricted $\mathbb{Z}_{11}^*$ multiplicative group (Figure \ref{fig:mod11_multiplication_no_zero_uqt}), the zero element is eliminated. Within this regime, every multiplication operation becomes a perfect bijection, a reversible cyclic permutation of the set. Restored to a purely unitary framework, the UQT exactly generalizes on the dataset, crystallizing into a deterministic 100\% accuracy without the stochastic instability characteristic of classical networks.

\textcolor{black}{The curves in Figure \ref{fig:mod11_plots} show a single run per task. Appendix~\ref{app:seeds} repeats the $\mathbb{Z}_{11}$ addition experiment with ten random seeds, all of which crystallize.} \textcolor{black}{A natural next question is what internal representation these crystallized models converge to. Appendix~\ref{app:group_structure} addresses this for modular addition as a representative case, and finds that the crystallized encoder organizes all 121 input pairs into 11 tight state clusters, one for each residue class.}

\subsection{Crystallization in non-Abelian algebra}

\begin{figure*}[tb!]
    \centering
    \subfloat[UQT: $S_4$ permutations\label{fig:s4_plots_uqt}]{%
        \includegraphics[width=0.48\textwidth]{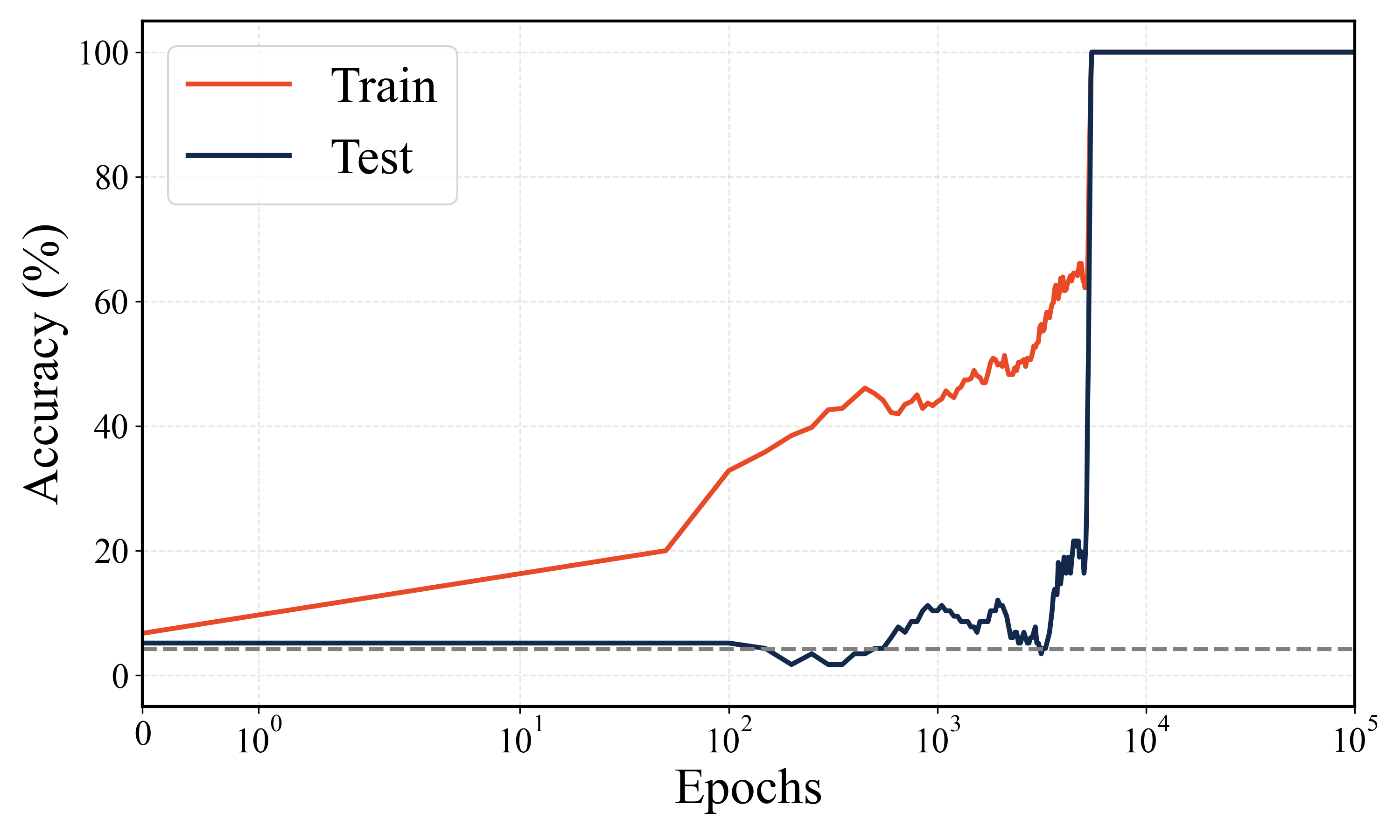}%
    }
    \hfill
    \subfloat[Transformer: $S_4$ permutations\label{fig:s4_plots_classical}]{%
        \includegraphics[width=0.48\textwidth]{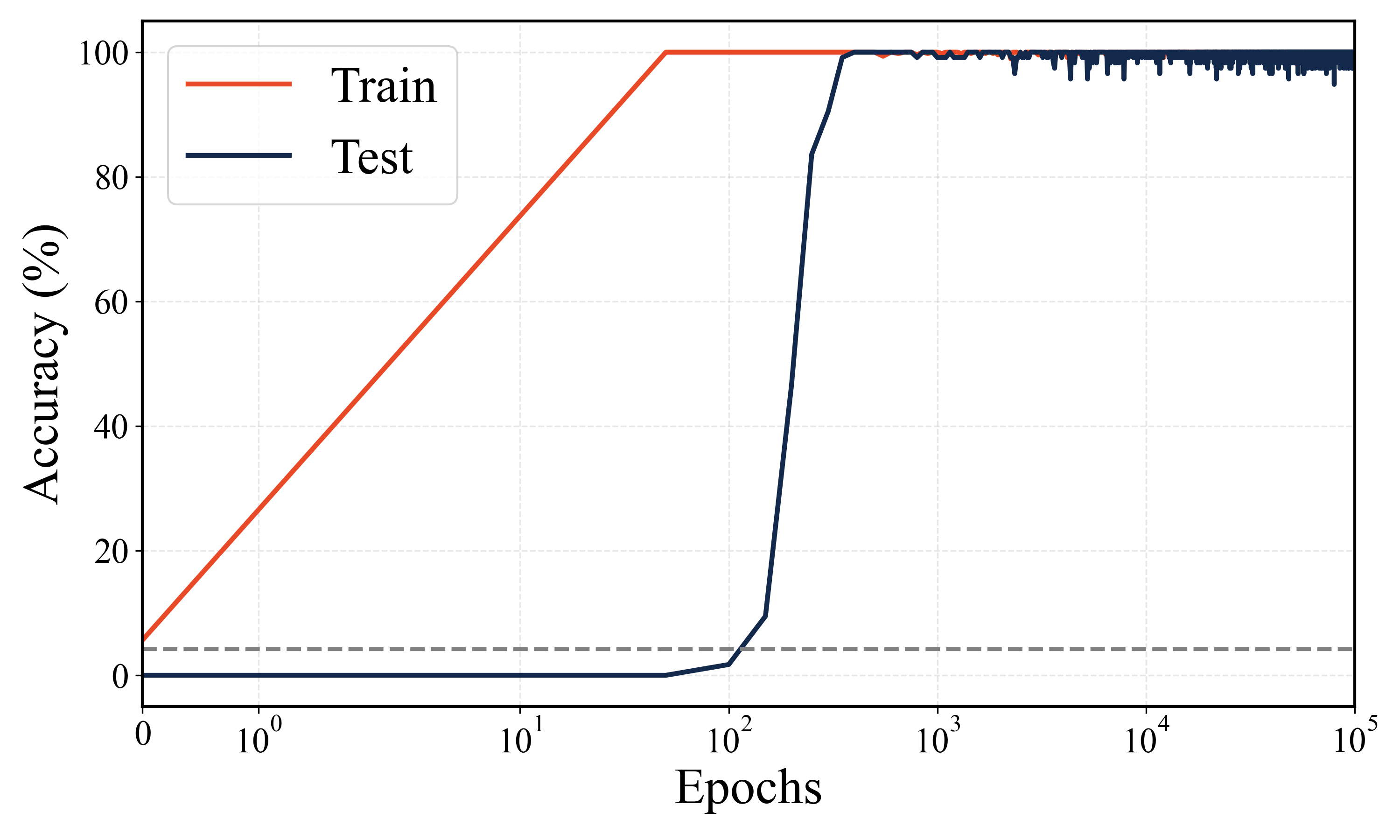}%
    }
    \caption{\textbf{Quantum crystallization in non-Abelian algebra.} \textcolor{black}{The UQT reaches and maintains 100\% test accuracy on the $S_4$ task, whereas the tested classical Transformer does not maintain stable 100\% generalization through the end of training.}}
    \label{fig:s4_plots}
\end{figure*}

While addition and multiplication are commutative, spatial and physical transformations are inherently non-commutative. To test this regime, we evaluated the architecture on the symmetric permutation group $S_4$. This abstract group consists of 24 distinct elements, meaning the network must learn its entire Cayley table. Here, a Cayley table is a strict algebraic multiplication table, defining the exact outcomes of all $24 \times 24 = 576$ possible paired combinations. \textcolor{black}{Because $S_4$ is non-Abelian (i.e., applying permutation $A$ followed by $B$ can yield a different result than $B$ followed by $A$), it provides a demanding test of whether a learned representation captures operation order. In our experiment, the classical Transformer eventually generalizes on this dataset but does not maintain stable 100\% test accuracy through the end of training (Figure \ref{fig:s4_plots_classical}).}

The quantum architecture, however, provides a native algebraic solution. The UQT embeds tokens using $SU(2)$ rotation matrices, the fundamental mathematical group that governs quantum spin and 3D spatial rotations. Because $SU(2)$ operations are intrinsically non-commutative, the physical wave-interference of the sequential quantum gates directly mirrors the geometric rules of the target algebra. By mapping the 24 abstract permutations directly into this continuous, non-commutative phase space, the UQT successfully crystallized across all 576 equations without statistical drift (Figure \ref{fig:s4_plots_uqt}). Note that to accommodate the higher representational density of the 24 permutations ($V=24$), we expanded the embeddings to the full register ($N_{emb}=5$) and reduced the mixing depth to $L=14$. Measuring the full 5-qubit register ($N=5$), this model requires a total of 690 parameters.

\subsection{Crystallization in linguistics}

\textcolor{black}{The aforementioned results establish that the UQT crystallizes across both commutative ($\mathbb{Z}_{11}$ addition and $\mathbb{Z}_{11}^{*}$ multiplication) and non-commutative ($S_4$) algebraic domains. To determine whether this advantage extends beyond periodic group operations to general rule learning, we turn to systematic linguistic compositionality. Unlike strict group operations, compositionality requires a model to independently isolate abstract conceptual dimensions (such as action and direction) and recombine them in configurations never encountered during training.} \textcolor{black}{The SCAN held-out ``jump'' split is a well-established challenge for systematic generalization in neural sequence models \cite{lake2018generalization}.}

\textcolor{black}{We evaluated this capacity using the ``add jump'' zero-shot split of the SCAN language \cite{lake2018generalization}. To strictly isolate the network's capacity for conceptual compositionality from the confounding dynamics of variable-length sequence generation, we distilled the core logic of this benchmark into a controlled, fixed-length dual-classification task. In our implementation, we mapped the $V=50$ vocabulary across the full $N_{emb}=N=6$ qubit register and utilized a mixing depth of $L=25$. Measuring the full 6-qubit register ($N=6$), this architecture requires only 1,650 trainable parameters.}

\textcolor{black}{To systematically test this, we constructed a vocabulary consisting of 10 optional prefix words (e.g., ``please,'' ``quickly''), 4 verb concepts (walk, run, jump, look) each containing 6 synonymous tokens, and 4 directional concepts (forward/none, left, right, around) each containing 4 synonymous tokens. Input sequences were strictly formulated as up to three tokens (\textit{Prefix, Verb, Direction}) padded to a fixed length of $S=3$. The objective is a dual-head classification task: the network must simultaneously predict the exact underlying verb concept (4 classes) and the directional concept (4 classes). To test true zero-shot composition, we partitioned the dataset such that the test set exclusively contains commands combining a ``jump'' concept with a directional modifier (e.g., ``jump left,'' ``spring backwards''). The training set contains all remaining combinations. Consequently, during training, the network sees the ``jump'' concept in isolation or with prefixes (e.g., ``quickly jump''), and it observes all directional modifiers applied to other verbs (e.g., ``walk left,'' ``look around''). However, it never observes ``jump'' combined with any direction (e.g., ``jump right'').}

\begin{figure*}[tb!]
    \centering
    \subfloat[UQT: SCAN ``add jump'' split\label{fig:scan_plots_uqt}]{%
        \includegraphics[width=0.48\textwidth]{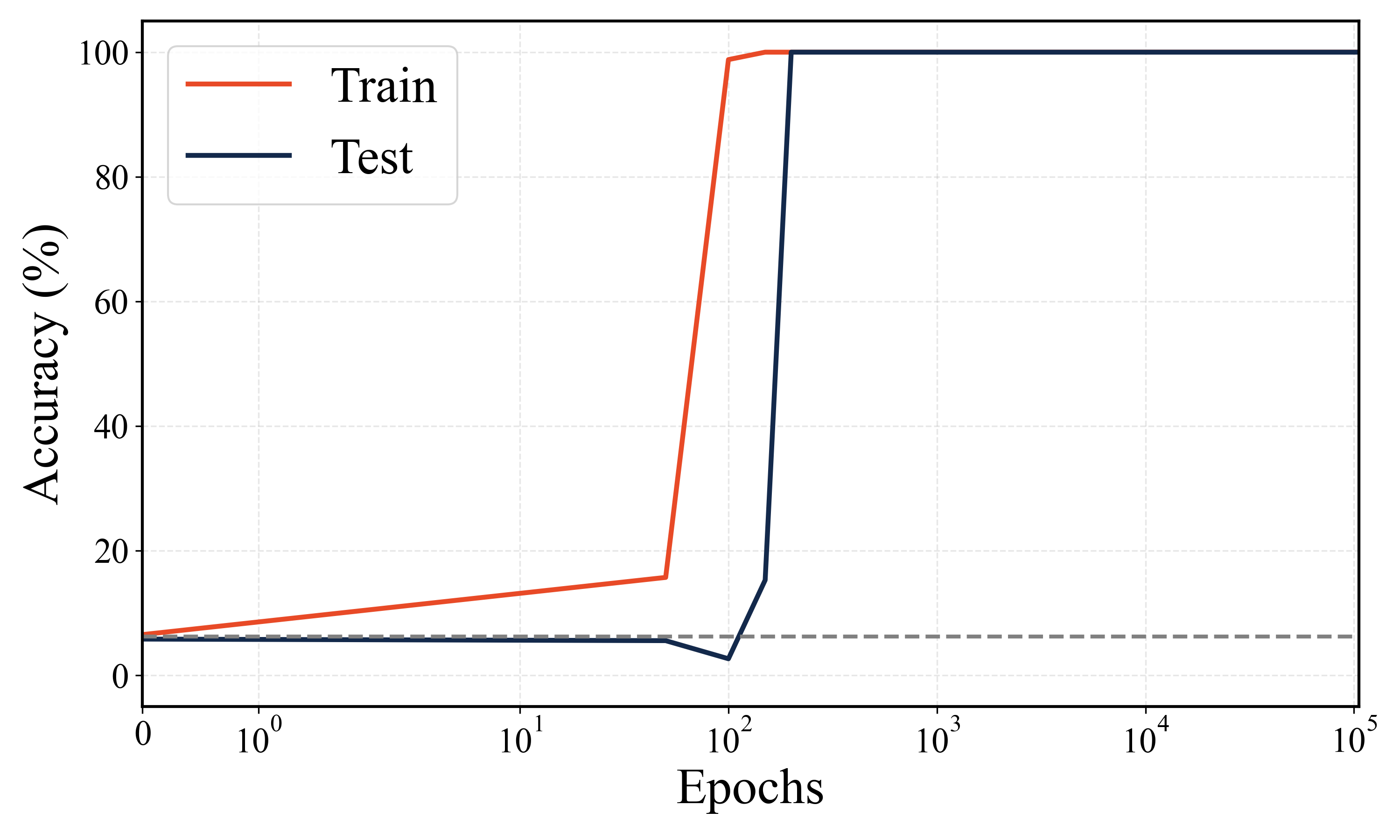}%
    }
    \hfill
    \subfloat[Transformer: SCAN ``add jump'' split\label{fig:scan_plots_classical}]{%
        \includegraphics[width=0.48\textwidth]{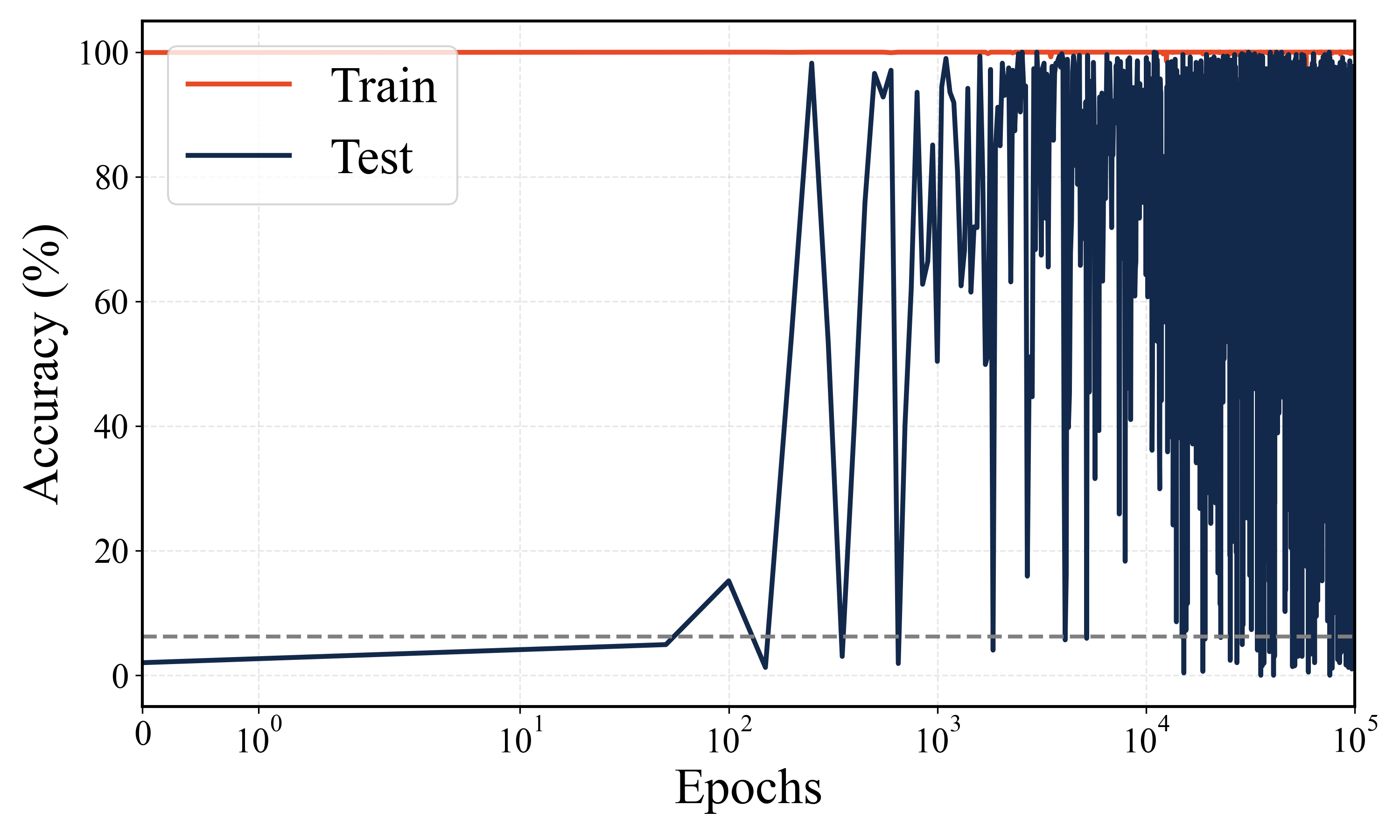}%
    }
    \caption{\textbf{Quantum crystallization in linguistic compositionality.} Zero-shot evaluation on the SCAN ``add jump'' split. \textcolor{black}{\textbf{(a)} The UQT reaches and maintains 100\% exact-match accuracy on unseen action--direction combinations. \textbf{(b)} The tested classical Transformer reaches 100\% training accuracy, but its held-out exact-match accuracy remains unstable through the end of training.}}
    \label{fig:scan_plots}
\end{figure*}

\textcolor{black}{Consistent with the difficulty previously reported for the held-out ``jump'' split \cite{lake2018generalization}, the classical Transformer evaluated here reaches 100\% training accuracy but does not maintain stable generalization on the zero-shot combinations. Its test accuracy fluctuates between 0\% and 100\% through the end of training (Figure \ref{fig:scan_plots_classical}).}

\textcolor{black}{In contrast, the UQT crystallized into exact, deterministic generalization. In our implementation, chronological linguistic tokens (prefixes, verbs, modifiers) are mapped directly to phase rotations, generating a superposed multi-qubit wave function. The architecture measures the resulting 64-state Hilbert space via two distinct conceptual heads: one identifying the verb construct (states 0--3) and the other identifying the modifier construct (states 4--7). Because the quantum circuit strictly obeys unitary $SU(2)$ boundaries, the geometric representations of action and direction do not destructively interfere or collapse into a statistical blend. Instead, they physically superimpose, allowing the network to organically decouple the action from the modifier. Leveraging this native physical compositionality, the quantum attention circuit zero-shot crystallizes, achieving stable 100\% exact-match accuracy across the unseen compositional combinations (Figure \ref{fig:scan_plots_uqt}).}

\subsection{Classical neural network baseline}

\textcolor{black}{A natural alternative explanation for the classical Transformer's inability to crystallize is its large parameter count (${\sim}$400,000 weights). To test whether parameter count alone accounts for the behavior of this baseline, we trained a compact standard classical MLP on all four experimental tasks where the UQT crystallizes.}

\textcolor{black}{The MLP architecture consists of a token embedding table followed by a single hidden layer with ReLU activation and a linear output projection. Embedding and hidden dimensions were chosen so that each task variant yields approximately 1,000 trainable parameters, comparable to the UQT parameter footprint across the four tasks (551--1,650).}

\begin{figure*}[tb!]
    \centering

    \subfloat[\textcolor{black}{MLP: Mod 11 addition ($\mathbb{Z}_{11}$)}\label{fig:nn_mod11_add}]{%
        \includegraphics[width=0.48\textwidth]{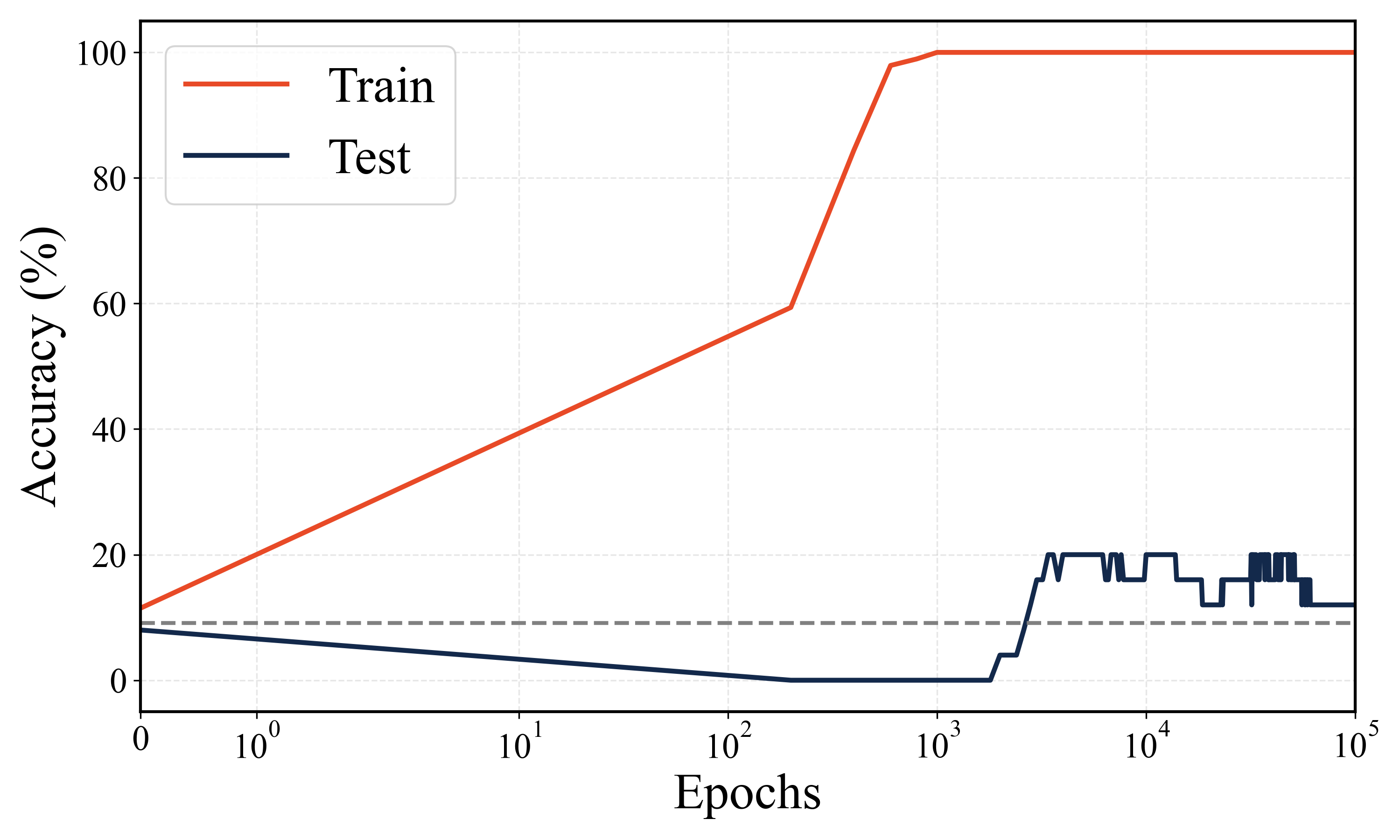}%
    }
    \hfill
    \subfloat[\textcolor{black}{MLP: Mod 11 multiplication ($\mathbb{Z}_{11}^*$)}\label{fig:nn_mod11_mul_nozero}]{%
        \includegraphics[width=0.48\textwidth]{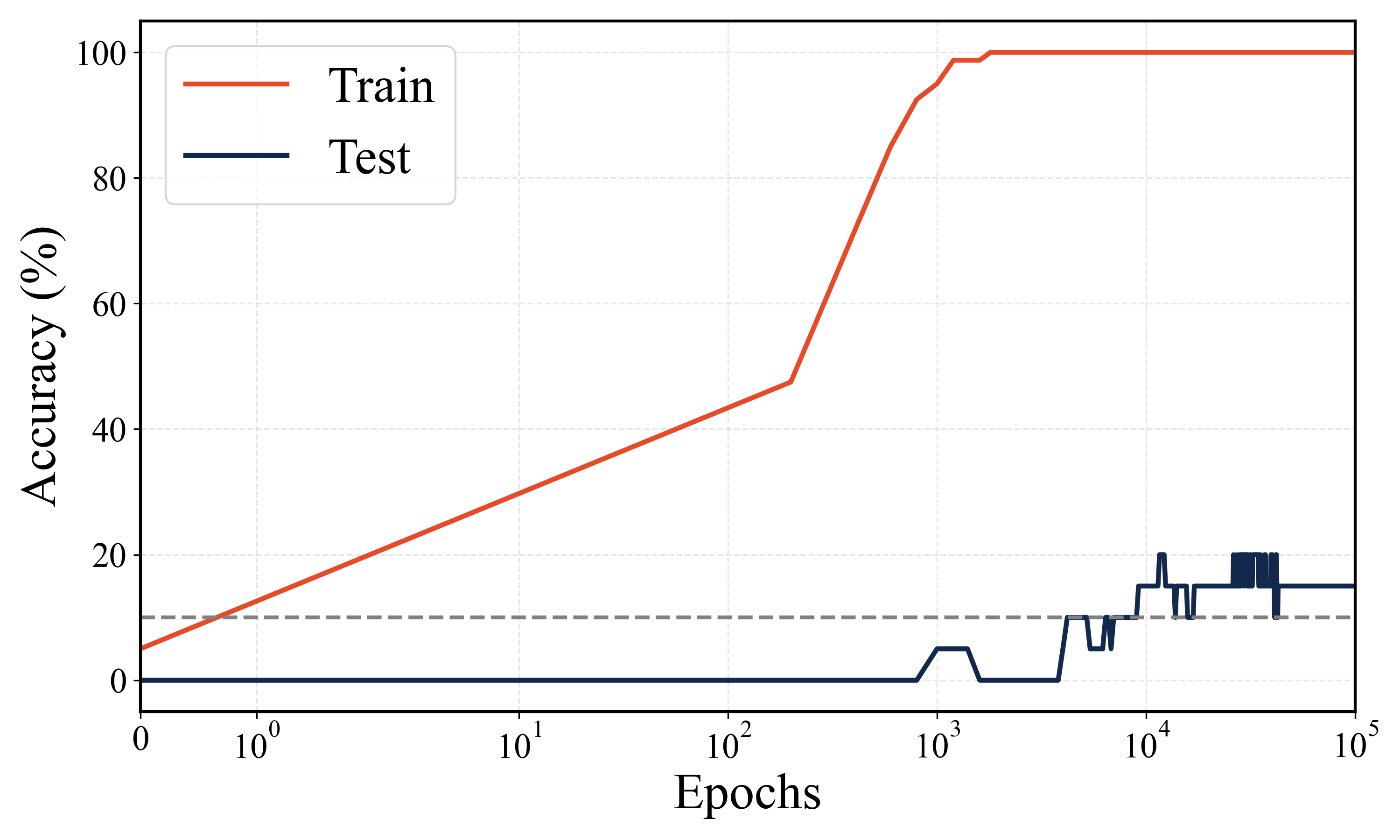}%
    }

    \vspace{0.4cm}

    \subfloat[\textcolor{black}{MLP: $S_4$ non-Abelian permutations}\label{fig:nn_s4}]{%
        \includegraphics[width=0.48\textwidth]{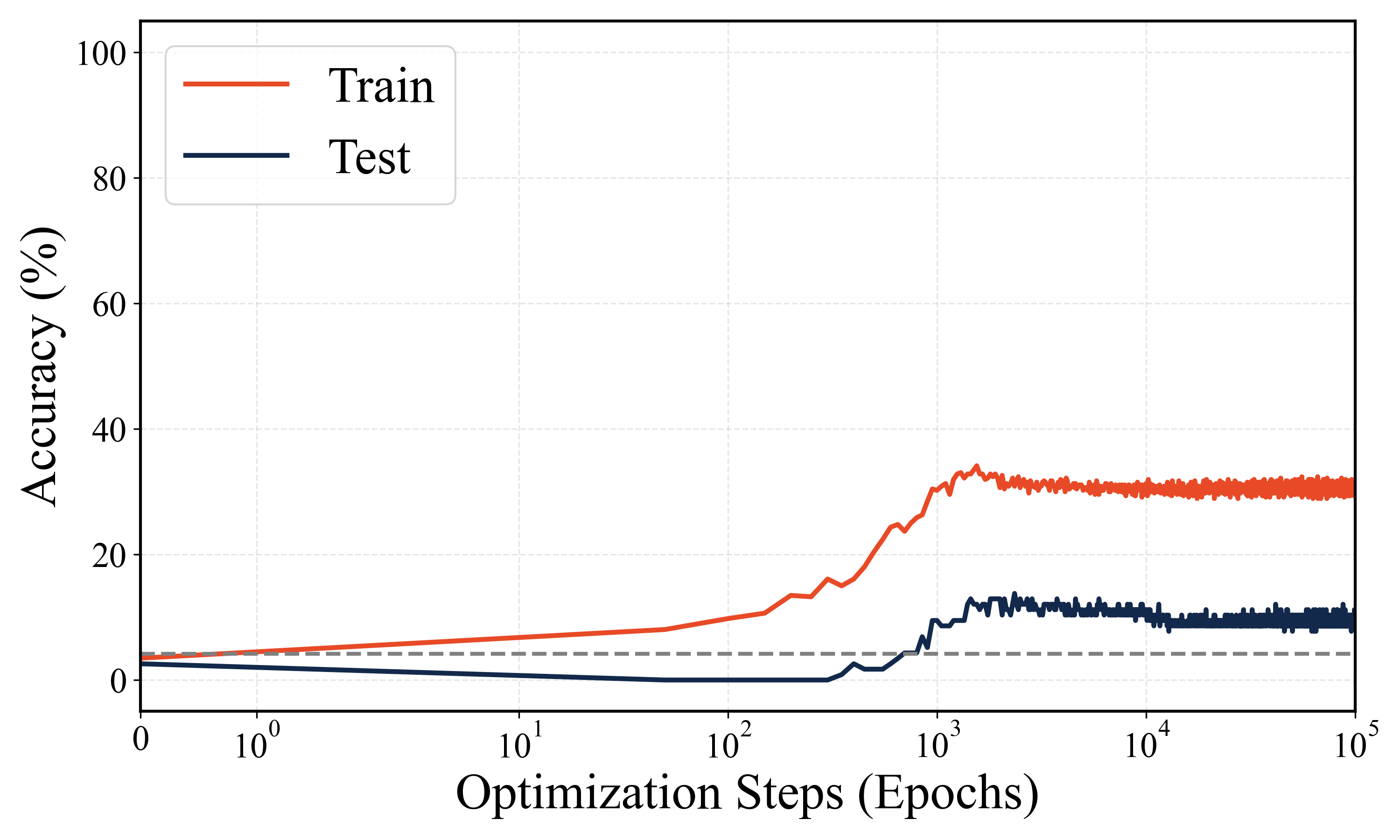}%
    }
    \hfill
    \subfloat[\textcolor{black}{MLP: SCAN ``add jump'' split}\label{fig:nn_scan}]{%
        \includegraphics[width=0.48\textwidth]{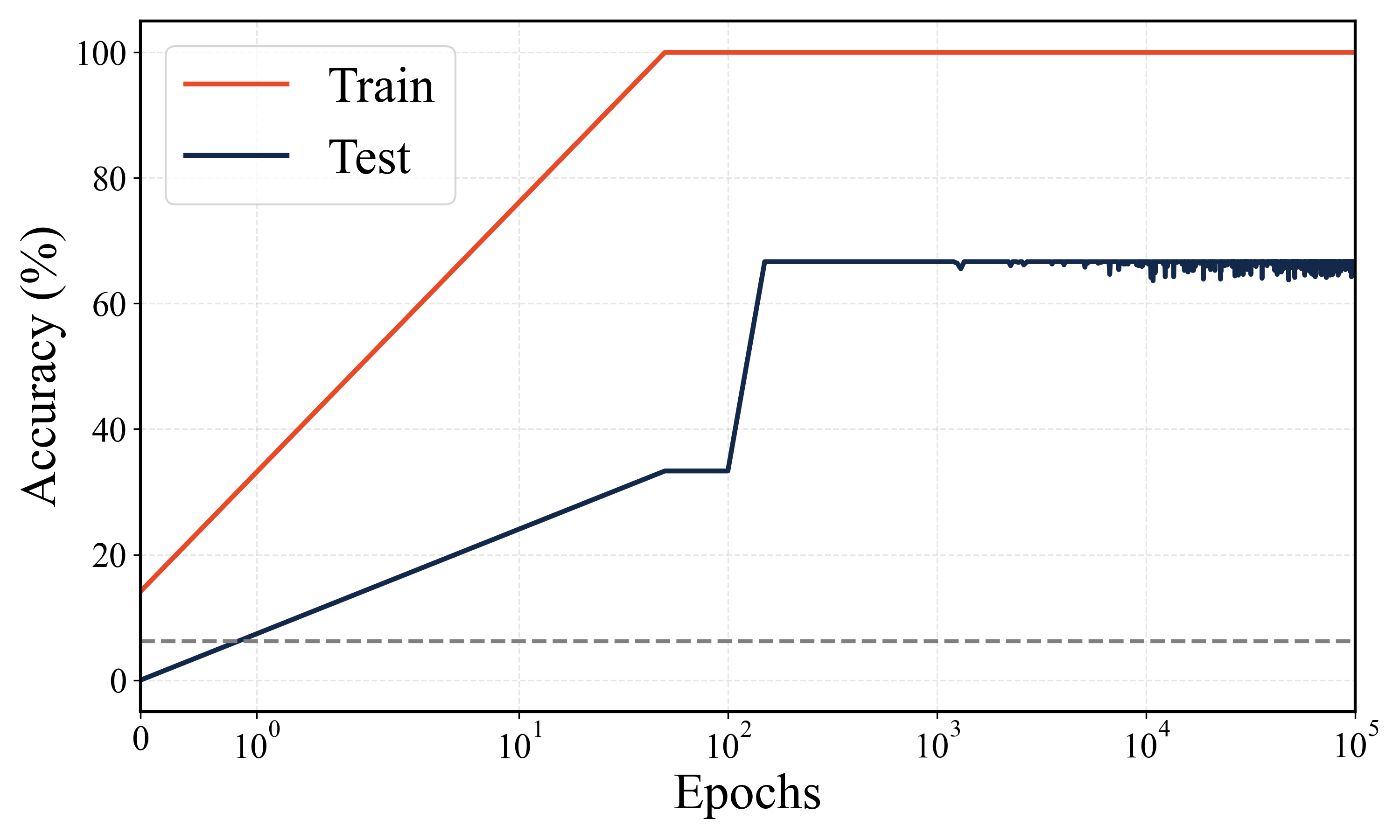}%
    }

    \caption{\textcolor{black}{\textbf{Classical MLP baseline: failure to crystallize across all four tasks.} A compact ${\sim}$1,000-parameter classical MLP trained with identical settings to the Transformer baseline. Across all four experimental domains where the UQT crystallizes, the standard MLP fails to reach 100\% test accuracy in any case, with no crystallization transition. Unlike the classical Transformer, which at least achieves delayed generalization via grokking, the MLP fails to grok.}}
    \label{fig:nn_baseline}
\end{figure*}

\textcolor{black}{As shown in Figure~\ref{fig:nn_baseline}, the compact MLP does not crystallize on any of the four tasks, and in no case does its test accuracy reach 100\%. Thus, under the evaluated architecture and training protocol, reducing parameter count alone does not resolve the behavior observed for the larger Transformer. A constructive analytical example of how quantum phase geometry can represent modular addition is provided in Appendix~\ref{app:theory}.}

\subsection{Physical deployment on NISQ hardware}
To empirically validate that the generalization capabilities of the UQT are not merely artifacts of noiseless classical simulation, we deployed the architecture onto physical superconducting quantum processors. In classical deep learning, gradients are computed via backpropagation. Physical quantum processors, however, lack an analytic computational graph, meaning exact analytic gradients must be evaluated empirically using the Parameter-Shift Rule \cite{mitarai2018quantum, schuld2019evaluating}. While the UQT's extreme parameter efficiency makes physical training theoretically viable, current NISQ devices suffer from two-qubit gate infidelities. The deep unitary depth required for structural entanglement ($L = 25$ layers in modular arithmetic and SCAN linguistic tasks, necessitating $>100$ sequential CNOT gates per forward pass) places unmitigated gradient-evaluation loops beyond the fidelity horizon of bare superconducting qubits.

\textcolor{black}{To bridge this gap, we adopted an offline inference protocol, consistent with recent approaches that validate classically trained quantum networks directly on physical hardware \cite{lakhdar2025benchmarking}. For each of the four experimental domains, the crystallized parameters $\vec{\theta}_{opt}$ obtained from the noiseless simulations described above were compiled into static physical quantum circuits, then evaluated on IBM Quantum hardware as a proof-of-concept physical validation. Full hardware specifications and execution details are provided in Appendix~\ref{app:hardware}.}

\begin{table*}[tb!]
\centering
\scriptsize{}
\renewcommand{\arraystretch}{0.88}
\caption{\textcolor{black}{\textbf{NISQ hardware inference results.} Unmitigated evaluation of the trained UQT parameters on physical IBM Quantum processors. The architecture achieved an overall 97.5\% success rate (39/40), including 100\% accuracy on all evaluated generalization test instances. Executions were distributed across three distinct hardware systems (\texttt{ibm\_marrakesh}, \texttt{ibm\_fez}, and \texttt{ibm\_kingston}) based on queue availability.}}
\label{tab:master_hardware}

\begin{tabular}{llcccc}
\toprule
\textbf{Domains} & \textbf{Input / Sequence} & \textbf{Target} & \textbf{IBM Output} & \textbf{Confidence} & \textbf{Status} \\
\midrule

\multicolumn{6}{l}{\textbf{Domain 1: $\mathbb{Z}_{11}$ Modular Addition (\texttt{ibm\_marrakesh})}} \\
\midrule
\textit{Train}
& $9 + 4 \pmod{11}$  & 2 & 2 & 22.0\% & \textbf{PASS} \\
\textit{(Memorization)} & $1 + 10 \pmod{11}$ & 0 & 0 & 22.4\% & \textbf{PASS} \\
& $2 + 7 \pmod{11}$  & 9 & 9 & 14.7\% & \textbf{PASS} \\
& $4 + 7 \pmod{11}$  & 0 & 0 & 24.1\% & \textbf{PASS} \\
& $6 + 6 \pmod{11}$  & 1 & 1 & 19.1\% & \textbf{PASS} \\
\midrule
\textit{Test}
& $5 + 0 \pmod{11}$  & 5 & 5 & 20.5\% & \textbf{PASS} \\
\textit{(Generalization)} & $3 + 3 \pmod{11}$  & 6 & 6 & 19.1\% & \textbf{PASS} \\
& $10 + 8 \pmod{11}$ & 7 & 7 & 17.6\% & \textbf{PASS} \\
& $5 + 10 \pmod{11}$ & 4 & 4 & 13.9\% & \textbf{PASS} \\
& $3 + 9 \pmod{11}$  & 1 & 1 & 19.4\% & \textbf{PASS} \\

\midrule
\multicolumn{6}{l}{\textbf{Domain 2: $\mathbb{Z}_{11}^*$ Modular Multiplication (\texttt{ibm\_fez})}} \\
\midrule
\textit{Train}
& $3 \times 5 \pmod{11}$ & 4 & 4 & 12.1\% & \textbf{PASS} \\
\textit{(Memorization)} & $6 \times 6 \pmod{11}$ & 3 & 3 & 11.5\% & \textbf{PASS} \\
& $1 \times 8 \pmod{11}$ & 8 & 8 & 10.8\% & \textbf{PASS} \\
& $1 \times 4 \pmod{11}$ & 4 & 4 & 8.5\%  & \textbf{PASS} \\
& $4 \times 6 \pmod{11}$ & 2 & 2 & 8.0\%  & \textbf{PASS} \\
\midrule
\textit{Test}
& $6 \times 4 \pmod{11}$ & 2 & 2 & 7.7\%  & \textbf{PASS} \\
\textit{(Generalization)} & $5 \times 5 \pmod{11}$ & 3 & 3 & 9.5\%  & \textbf{PASS} \\
& $4 \times 10 \pmod{11}$& 7 & 7 & 13.1\% & \textbf{PASS} \\
& $5 \times 6 \pmod{11}$ & 8 & 8 & 10.4\% & \textbf{PASS} \\
& $2 \times 9 \pmod{11}$ & 7 & 7 & 7.8\%  & \textbf{PASS} \\

\midrule
\multicolumn{6}{l}{\textbf{Domain 3: $S_4$ Non-Abelian Permutations (\texttt{ibm\_marrakesh})}} \\
\midrule
\textit{Train}
& $P(3) \circ P(22)$  & 10 & 10 & 9.7\%  & \textbf{PASS} \\
\textit{(Memorization)} & $P(15) \circ P(17)$ & 18 & 13 & 6.3\%  & \textbf{FAIL*} \\
& $P(19) \circ P(1)$  & 18 & 18 & 14.2\% & \textbf{PASS} \\
& $P(7) \circ P(23)$  & 16 & 16 & 9.1\%  & \textbf{PASS} \\
& $P(6) \circ P(13)$  & 15 & 15 & 13.5\% & \textbf{PASS} \\
\midrule
\textit{Test}
& $P(9) \circ P(19)$  & 1  & 1  & 13.8\% & \textbf{PASS} \\
\textit{(Generalization)} & $P(23) \circ P(11)$ & 12 & 12 & 9.5\%  & \textbf{PASS} \\
& $P(4) \circ P(13)$  & 6  & 6  & 10.6\% & \textbf{PASS} \\
& $P(3) \circ P(5)$   & 1  & 1  & 9.9\%  & \textbf{PASS} \\
& $P(18) \circ P(8)$  & 1  & 1  & 9.3\%  & \textbf{PASS} \\

\midrule
\multicolumn{6}{l}{\textbf{Domain 4: SCAN Linguistic Compositionality (\texttt{ibm\_kingston})}} \\
\midrule
\textit{Train}
& now saunter ahead      & Walk, None   & Walk, None   & 44.4\% / 41.1\% & \textbf{PASS} \\
\textit{(Memorization)} & carefully stroll back  & Walk, Around & Walk, Around & 45.5\% / 47.8\% & \textbf{PASS} \\
& suddenly race westward & Run, Left    & Run, Left    & 41.5\% / 42.6\% & \textbf{PASS} \\
& suddenly walk left     & Walk, Left   & Walk, Left   & 47.4\% / 42.5\% & \textbf{PASS} \\
& now glance back        & Look, Around & Look, Around & 38.7\% / 40.1\% & \textbf{PASS} \\
\midrule
\textit{Test}
& slowly jump port       & Jump, Left   & Jump, Left   & 42.1\% / 44.9\% & \textbf{PASS} \\
\textit{(Zero-Shot)} & now leap right         & Jump, Right  & Jump, Right  & 40.7\% / 42.4\% & \textbf{PASS} \\
& slowly jump right      & Jump, Right  & Jump, Right  & 37.4\% / 44.2\% & \textbf{PASS} \\
& hop behind             & Jump, Around & Jump, Around & 34.6\% / 35.7\% & \textbf{PASS} \\
& quietly hop westward   & Jump, Left   & Jump, Left   & 36.8\% / 36.9\% & \textbf{PASS} \\
\bottomrule
\end{tabular}

\vspace{0.15cm}
\parbox{0.97\textwidth}{\scriptsize \textit{*Note: The $S_4$ task experienced a single physical training-set failure. However, raw readout analysis confirmed the target remained the distinct second-most probable state (4.7\%), retaining a clear probability peak above the 3.1\% physical random noise floor.}}
\end{table*}

As detailed in Table \ref{tab:master_hardware}, the quantum architecture demonstrated robust physical resilience. Across \textcolor{black}{40} unmitigated hardware evaluations, the UQT achieved an overall physical success rate of \textcolor{black}{97.5\% (39/40)}. It is crucial to interpret the raw confidence values through the lens of quantum measurement dimensionality. The Hilbert space for \textcolor{black}{the algebraic and arithmetic evaluations on 5-qubit registers} encompasses $2^5 = 32$ basis states, resulting in a random physical noise floor of $\sim 3.1\%$. The recorded confidence peaks across the passing evaluations ($7\%$ to $24\%$) represent significant, statistically decisive constructive interference overriding environmental decoherence. \textcolor{black}{For the SCAN linguistic task, the architecture utilizes a 6-qubit register ($2^6 = 64$ basis states) and measures two independent 4-state sub-spaces (verb and modifier). Within each classification head, the normalized physical random noise floor is exactly 25\%. The network consistently generated dominant probability peaks ($34\%$ to $47\%$) simultaneously across both semantic heads. By decisively overcoming hardware decoherence in both sub-spaces, the architecture easily bypassed the rigorous 6.25\% ($1/4 \times 1/4$) random exact-match baseline, maintaining a perfect 100\% zero-shot exact-match accuracy.}

\begin{figure*}[tb!]
    \centering
    \subfloat[JAX simulation result\label{fig:hw_ideal}]{%
        \includegraphics[width=0.75\textwidth]{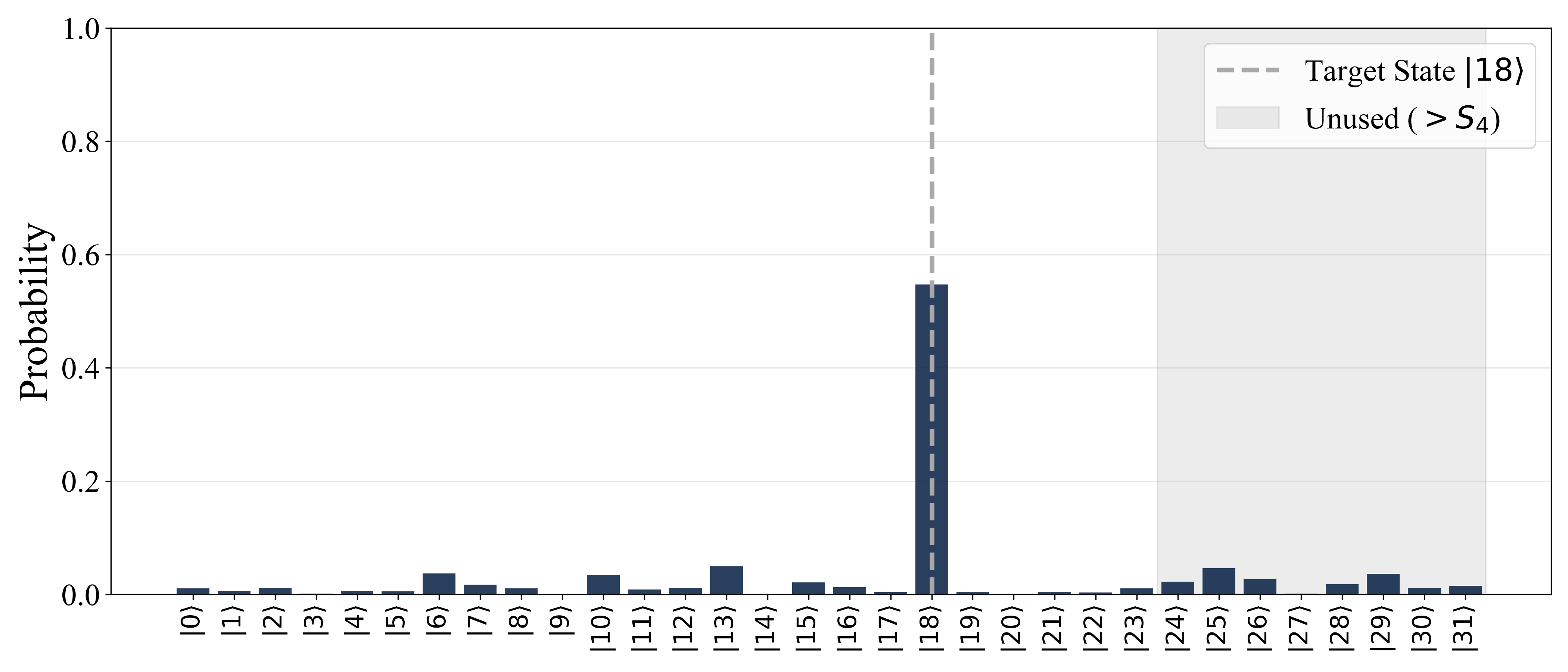}%
    }

    \vspace{0.5cm}

    \subfloat[Hardware inference result\label{fig:hw_ibm}]{%
        \includegraphics[width=0.75\textwidth]{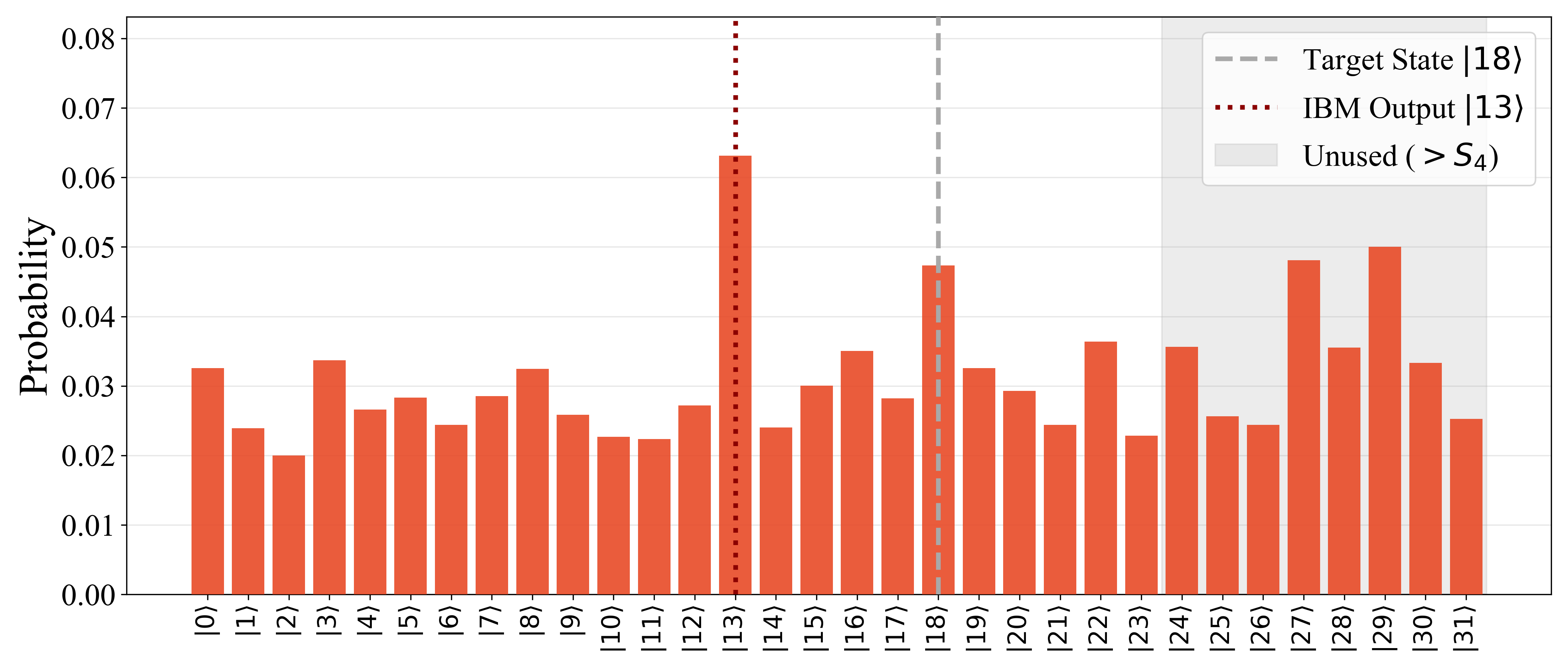}%
    }
    \caption{\textbf{Quantum state amplitude distribution for the failed $S_4$ permutation $P(15) \circ P(17)$.} \textbf{(a)} Ideal probability distribution obtained via noiseless JAX simulation, isolating the target state ($|18\rangle$). \textbf{(b)} Raw probability distribution evaluated on the \texttt{ibm\_marrakesh} physical processor. \textcolor{black}{Under unmitigated hardware noise, state $|13\rangle$ produced the largest observed probability (6.3\%), while the true target state $|18\rangle$ remained the distinct runner-up (4.7\%), above the 3.1\% uniform random baseline.}}
    \label{fig:histogram}
\end{figure*}

The single misclassification occurred within the training set of the highly complex $S_4$ non-Abelian domain, where the network erroneously predicted Class 13 instead of the target Class 18. However, analysis of the raw probability distribution generated across 8,192 shots (Figure \ref{fig:histogram}) revealed that the true target remained the distinct second-most probable state (4.7\%), successfully maintaining a probability peak above the 3.1\% random noise floor. This specific failure mode illustrates the boundary of current NISQ fidelity; while the geometric wave-interference correctly isolated the target amplitude within the simulated model, unmitigated hardware decoherence allowed an erroneous state ($|13\rangle$) to marginally overtake the primary signal (6.3\%). Despite this, the 100\% success rate across all evaluated generalization test instances confirms that the UQT's geometric parameterization inherently generates highly stable physical wave-interference patterns capable of surviving modern NISQ constraints.

\subsection{Asymptotic complexity}

\noindent Consider the representational scaling required to model exact formal systems. In standard classical Transformers, embedding a finite token space of size $V$ requires mapping tokens into a high-dimensional Euclidean space, with an embedding table demanding $\mathcal{O}(V \times d_{\text{model}})$ parameters. Furthermore, classical multi-head attention and feed-forward networks rely on dense projection matrices that scale quadratically with this hidden dimension, imposing a massive $\mathcal{O}(d_{\text{model}}^2)$ parameter bottleneck. Conversely, the UQT maps tokens directly into the continuous phase amplitudes of an $N$-qubit register. Because the multi-qubit Hilbert space scales exponentially ($\mathcal{O}(2^N)$ state dimensions), the UQT achieves massive representational density using only a logarithmic number of physical qubits: $\mathcal{O}(\log V)$. Consequently, the quantum embedding dimension is strictly bounded by the number of phase angles applied to the $N$ qubits, reducing the representation scaling to $\mathcal{O}(V \times \log V)$, while the sequential quantum mixing layers require only $\mathcal{O}(L \times \log V)$ parameters. This allows the UQT to bypass the $\mathcal{O}(d_{\text{model}}^2)$ classical parameter explosion entirely.

Crucially, this logarithmic spatial scaling reveals a profound connection to foundational quantum algorithms. In our modular arithmetic evaluations, the UQT natively learns to decode cyclic, periodic group structures, mathematics fundamentally akin to the phase-estimation logic required for the modular exponentiation step in Shor's algorithm \cite{shor1999polynomial}. However, rather than relying on a rigid, general-purpose Inverse Quantum Fourier Transform (IQFT) to extract the phase, the parameterized mixing layers organically learn a highly compressed, task-specific basis transformation. This demonstrates that the UQT provides \textcolor{black}{a parameter-efficiency advantage} for learning the exact discrete formal systems that currently \textcolor{black}{force} classical continuous-space models into massive over-parameterization.

\section{Discussion}
\textcolor{black}{Using an identical quantum attention topology with 5 or 6 qubits and 551--1,650 trainable parameters, the UQT crystallizes on four tasks spanning cyclic modular arithmetic, non-Abelian group algebra, and systematic linguistic compositionality.} \textcolor{black}{Physical deployment on IBM Quantum hardware further demonstrates that the learned parameters transfer beyond noiseless simulation,} \textcolor{black}{correctly classifying 39 of 40 selected instances under unmitigated NISQ noise.}

\textcolor{black}{Neither the tested MLP nor the tested Transformer crystallizes on these tasks under the reported protocols. This comparison indicates that the UQT provides a useful structural inductive bias for exact formal reasoning relative to the classical architectures evaluated in this paper.}

\textcolor{black}{{Notably, the present experiments operate at a relatively small scale. Extending this architecture to substantially larger formal domains remains an open challenge. Nevertheless, the UQT's compact parameter footprint establishes a structural foundation for addressing this as quantum hardware matures.}}

\section*{Author contributions}
S.C. conceived the Universal Quantum Transformer architecture and designed its quantum circuit topologies, implemented the JAX and PyTorch codebases, executed the inference evaluations on physical IBM Quantum hardware, and wrote the original manuscript. A.T. supervised the research, collaborated in advancing and refining the initial version of the model architecture, contributed to the theoretical analysis, validated the algorithms and code, and critically revised the writing.

\section*{Funding Declaration}
The authors declare that they have no funding sources to disclose for this work.

\appendix
\section{Mathematical formulation of the UQT}
\label{app:math}
The core function of the UQT relies on encoding classical tokens into geometric quantum states. For an input token $x$ mapped to a parameterized vector $\vec{\theta}_x$, the embedding unitary $\mathcal{E}$ is applied across a dedicated $N_{emb}$-qubit subspace of the total $N$-qubit register:
\begin{align}
    \mathcal{E}(\vec{\theta}_x) &= \Biggl( \bigotimes_{q=0}^{N_{emb}-1} R_y(\theta_{x,q,3}) R_x(\theta_{x,q,2}) \nonumber \\
    &\quad R_y(\theta_{x,q,1}) R_x(\theta_{x,q,0}) \Biggr) \otimes I^{\otimes (N - N_{emb})},
\end{align}
where $I$ represents the identity operation on the remaining un-embedded ancilla qubits. This alternating sequence of orthogonal rotations provides universal single-qubit control, ensuring that each discrete token can be mapped to any arbitrary geometric coordinate on the Bloch sphere. In our empirical evaluations, we embedded operands into an $N_{emb}=4$ qubit subspace on a 5-qubit register ($N=5$) for modular arithmetic, and utilized the full 5-qubit register ($N_{emb}=N=5$) for the non-Abelian permutations. \textcolor{black}{For the linguistic compositionality task, all 6 qubits were used as embedding qubits ($N_{emb}=N=6$).}

Following the rotational embeddings, the quantum state is processed by digital mixing layers to generate deep structural entanglement. Each layer $l$ applies a parameterized general $SU(2)$ rotation to every qubit, followed by a global cyclic ring of CNOT gates ($C_{\mathrm{ring}}$):
\begin{equation}
    U_{\mathrm{mix}}^{(l)} = C_{\mathrm{ring}} \bigotimes_{q=0}^{N-1} U_{\mathrm{Rot}}^{(l)}\left(\alpha_{l,q}, \beta_{l,q}, \gamma_{l,q}\right).
\end{equation}

In the quantum attention circuit, a full set of tokens $T = (x_1, x_2, \dots, x_S)$ are embedded consecutively prior to any entanglement, allowing their geometric states to natively superimpose. This state is subsequently processed by $L$ mixing layers to produce the final pre-measurement state:
\begin{equation}
    |\psi_{attn}\rangle = \left( \prod_{l=1}^{L} U_{\mathrm{mix}}^{(l)} \right) \left( \prod_{t=1}^{S} \mathcal{E}(\vec{\theta}_{x_t}) \right) |0\rangle^{\otimes N},
\end{equation}
where the embedding product denotes time-ordered application from right to left.

\section{Implementation details}
\label{app:impl}
\subsection{Optimization and asymmetric regularization}
The architecture was optimized using the JAX high-performance array computing framework \cite{jax2018github} and the Optax AdamW optimizer. Across all experimental domains, we utilized a constant learning rate of $\eta = 0.005$ alongside moment decay rates of $\beta_1 = 0.9$ and $\beta_2 = 0.98$. We applied an asymmetric loss function utilizing an $L_2$ weight decay ($\lambda_{ent} = 0.01$) exclusively on the entanglement logic gates ($W_{ent}$), while the phase embedding angles ($W_{emb}$) were masked from penalization:
\begin{equation}
    L_{total} = \mathcal{L}_{\text{NLL}} + \lambda_{ent} ||W_{ent}||_2^2,
\end{equation}
where $W_{ent}$ encompasses the rotational parameters of the mixing layers, and $W_{emb}$ represents the unpenalized token embedding phases. This architectural constraint permits the token geometry to rotate freely across the Bloch sphere. Simultaneously, the $L_2$ penalty forces unnecessary entanglement parameters toward zero, functionally reducing arbitrary $SU(2)$ rotations into Identity operations ($I$). This dynamic natively prunes the circuit during optimization, severely restricting the logical density of the CNOT routing network and preventing over-entanglement.

\subsection{Dataset generation}

Datasets for modular arithmetic over $\mathbb{Z}_{11}$ and $\mathbb{Z}_{11}^*$, as well as $S_4$ permutations, were partitioned into 80/20 train/test splits. \textcolor{black}{The SCAN linguistic task utilized a strict zero-shot compositional split, where the test set exclusively contains commands combining the ``jump'' concept with directional modifiers unseen in training combinations. The full vocabulary ($V=50$) consists of: 10 optional prefix tokens (\textit{please, kindly, just, quickly, now, carefully, quietly, slowly, simply, suddenly}), with no-prefix also permitted; 4 verb concepts each represented by 6 synonymous surface tokens (Walk: \textit{walk, stroll, amble, saunter, march, pace}; Run: \textit{run, sprint, dash, jog, race, rush}; Jump: \textit{jump, leap, hop, bound, vault, spring}; Look: \textit{look, stare, gaze, glance, peek, peer}); and 4 directional concepts each represented by up to 4 surface tokens (Forward/None: \textit{no modifier, straight, forward, ahead}; Left: \textit{left, port, counterclockwise, westward}; Right: \textit{right, starboard, clockwise, eastward}; Around: \textit{around, back, backwards, behind}).}

\textcolor{black}{Because all four task domains are finite, the held-out test sets constitute exhaustive evaluations: the 20\% partitions cover all held-out operand pairs for modular arithmetic and $S_4$, and the zero-shot split covers all possible compositions of the jump concept with directional modifiers in the SCAN task.}

\subsection{Classical Transformer baseline}

To establish a rigorous classical baseline, we trained standard attention-based Transformers using PyTorch \cite{paszke2019pytorch}. The classical architecture comprised 2 encoder layers, 4 attention heads, an embedding dimension of $d_{\text{model}}=128$, and a feedforward network dimension of 512, requiring approximately $4 \times 10^5$ trainable parameters. \textcolor{black}{This architecture provides a substantially higher-capacity classical comparator to the UQT's sub-2,000-parameter configurations.} \textcolor{black}{Training followed the optimization protocol of Power et al.\ \cite{power2022grokking}: AdamW optimizer with $\eta{=}10^{-3}$, $\beta_1{=}0.9$, $\beta_2{=}0.98$, weight decay $\lambda{=}1.0$, and a linear warmup scheduler over the first 10 updates. A minibatch size of 48 was used for all arithmetic and permutation tasks, and 64 for the SCAN linguistic task; these batch sizes were held fixed across all three model types (UQT, Transformer, and MLP). }

\subsection{Classical MLP baseline}
\label{app:mlp}

\textcolor{black}{To test whether matching the UQT's parameter scale changes the classical-baseline behavior, we additionally trained a compact classical MLP on all four tasks where the UQT crystallizes.} \textcolor{black}{The architecture follows: $\text{Embedding}(V, d) \rightarrow \text{flatten} \rightarrow \text{Linear}(S \cdot d,\, h) \rightarrow \text{ReLU} \rightarrow \text{Linear}(h,\, C)$, where $V$ is vocabulary size, $d$ the embedding dimension, $S$ the sequence length, $h$ the hidden dimension, and $C$ the number of output classes. For the SCAN task, the final layer is replaced by two parallel output heads (verb and modifier), identical in structure to the classical Transformer baseline. Parameter counts were chosen to be comparable to the UQT: 995 parameters for modular arithmetic tasks ($\mathbb{Z}_{11}$ addition and $\mathbb{Z}_{11}^*$ multiplication, $d{=}8$, $h{=}32$), 984 for $S_4$ ($d{=}5$, $h{=}24$), and 956 for SCAN ($d{=}6$, $h{=}24$). All training hyperparameters were held fixed and identical to the classical Transformer baseline.}

\section{Seed robustness}
\label{app:seeds}

\textcolor{black}{We repeated the $\mathbb{Z}_{11}$ addition experiment under the identical protocol with ten random seeds, each setting the data split, the initialization, and the minibatch order together. All ten reached 100\% test accuracy and held it through the end of training at 20,000 epochs. The onset of crystallization ranged from epoch 650 to epoch 1800, with a median of 1150, and the run in Figure~\ref{fig:mod11_addition_uqt} crystallized at epoch 1250.}

\begin{figure}[tb!]
    \centering
    \includegraphics[width=\columnwidth]{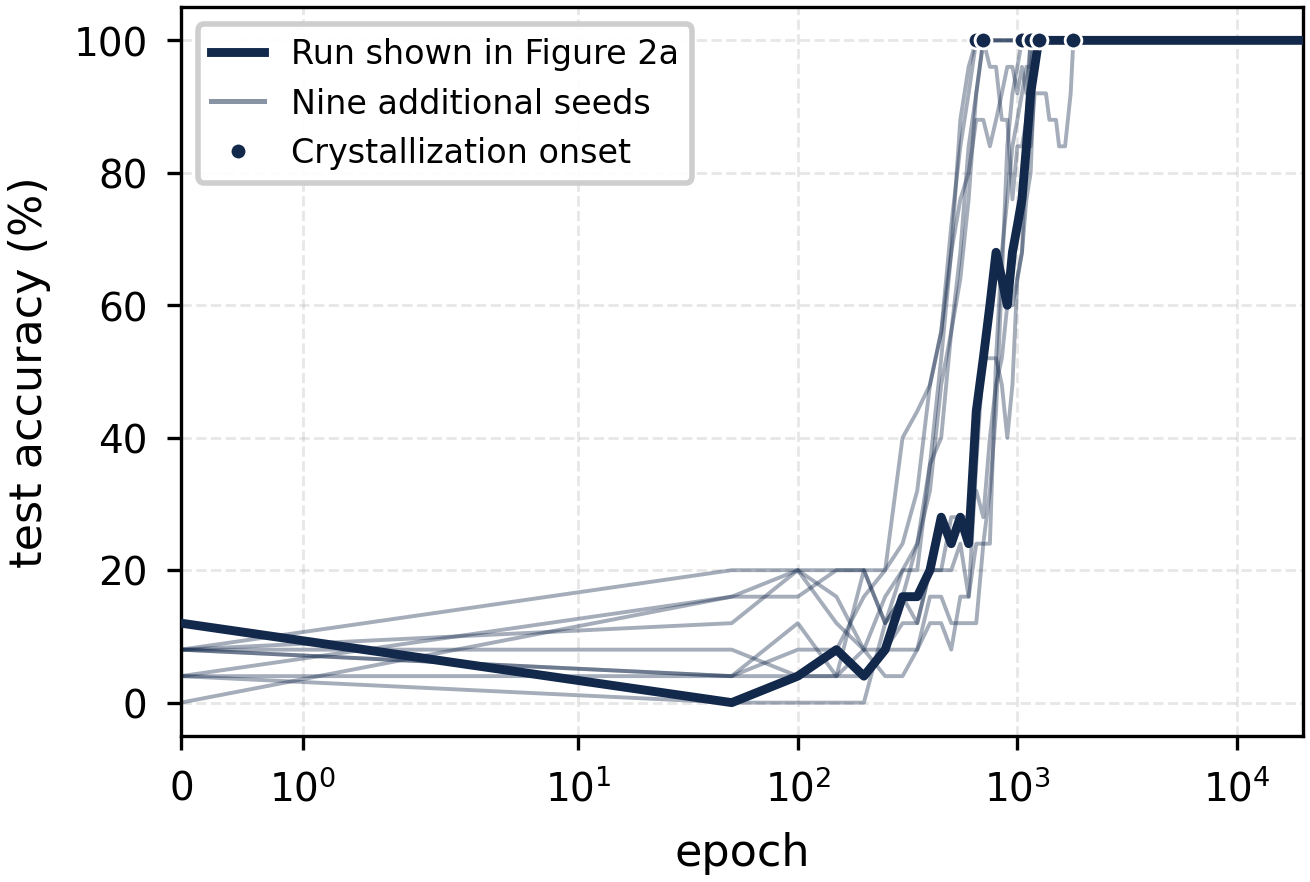}
    \caption{\textcolor{black}{\textbf{Crystallization across ten seeds ($p=11$ modular addition).} Test accuracy for ten runs differing only in the random seed. Markers give the crystallization onset $t_c$, after which every recorded evaluation stays at 100\% through epoch 20,000.}}
    \label{fig:seed_robustness}
\end{figure}

\section{Emergent group structure in the crystallized encoder}
\label{app:group_structure}

\begin{figure*}[tb!]
    \centering
    \subfloat[Initialization\label{fig:group_collapse_init}]{%
        \includegraphics[width=0.32\textwidth]{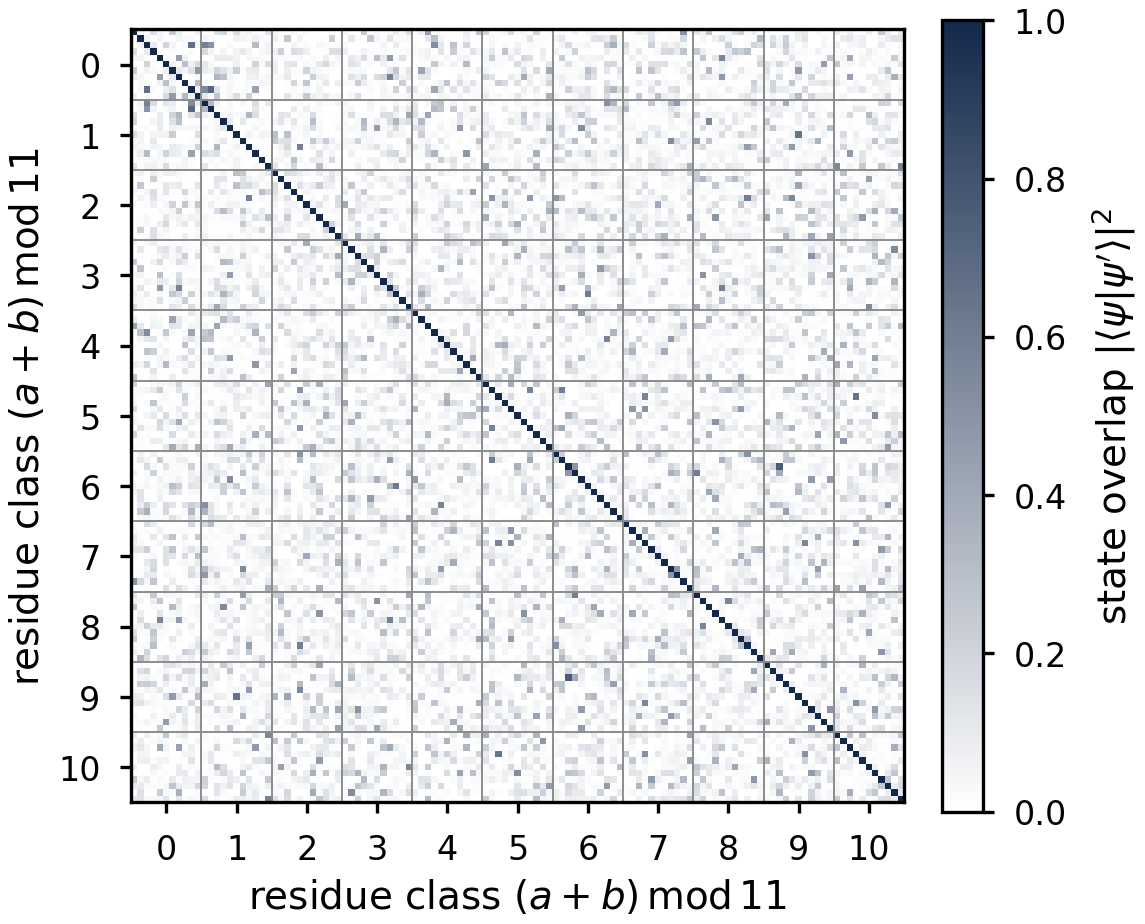}%
    }
    \hfill
    \subfloat[After crystallization\label{fig:group_collapse_trained}]{%
        \includegraphics[width=0.32\textwidth]{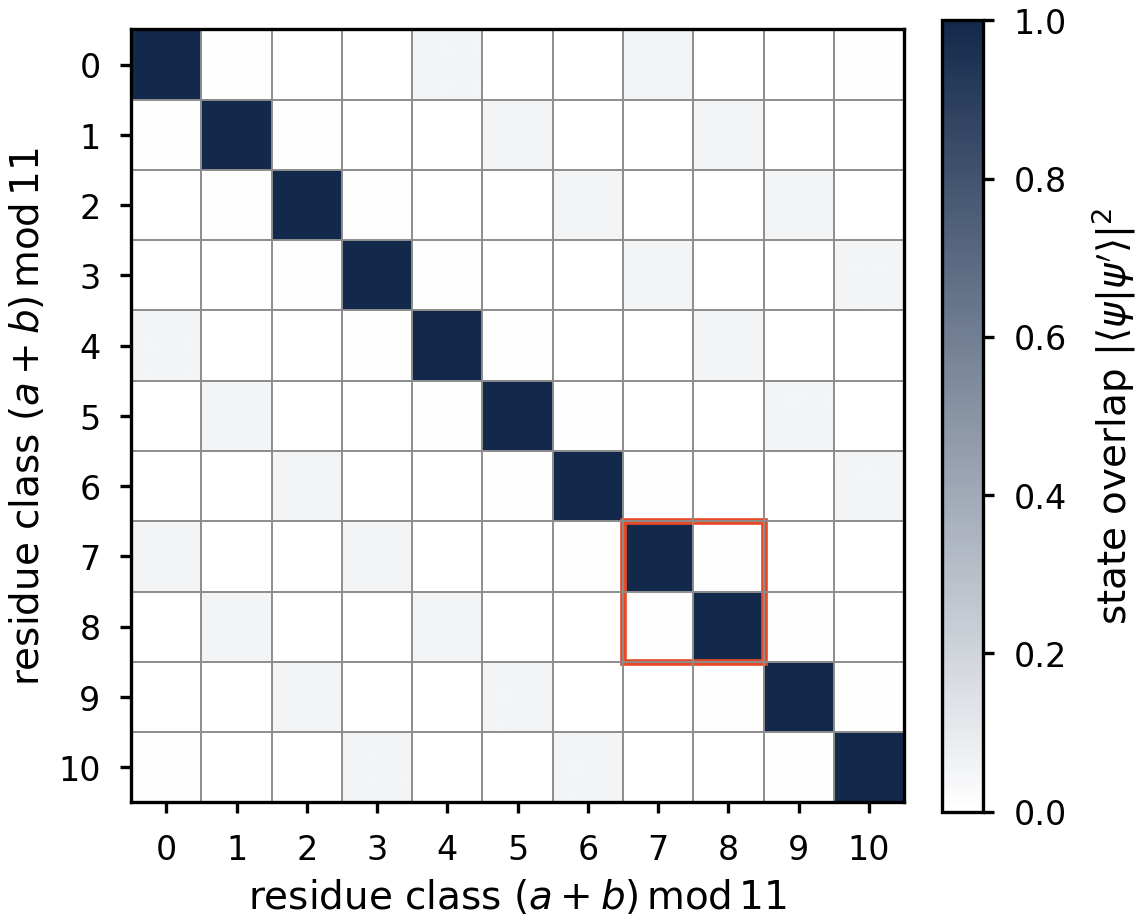}%
    }
    \hfill
    \subfloat[Magnification of classes 7 and 8\label{fig:group_collapse_zoom}]{%
        \includegraphics[width=0.307\textwidth]{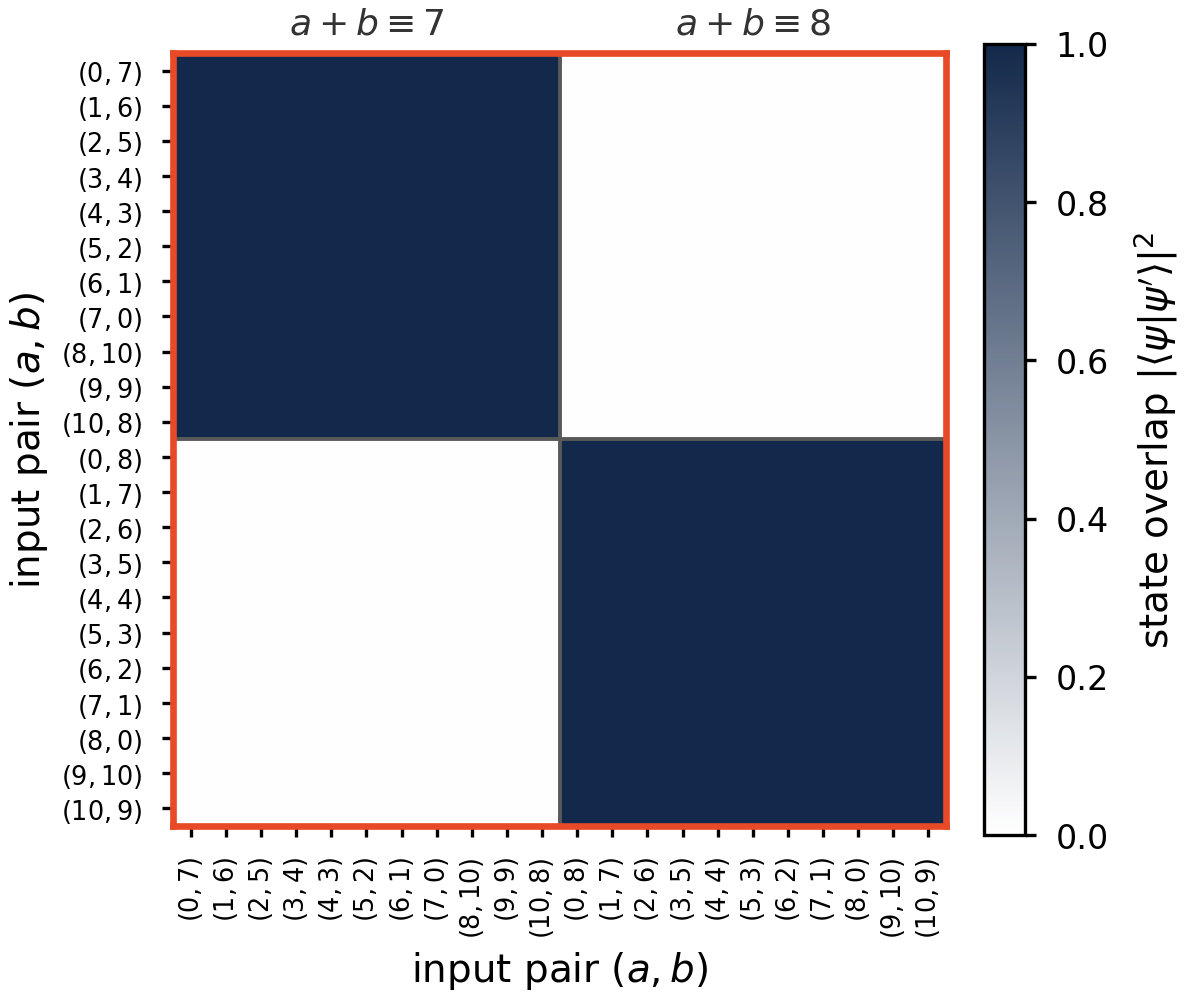}%
    }
    \caption{\textcolor{black}{\textbf{Emergent cyclic group structure ($p=11$ modular addition).} Overlaps $|\langle\psi(a,b)|\psi(a',b')\rangle|^2$ between the states prepared by all 121 ordered input pairs, grouped by residue class $(a+b)\bmod 11$, so each block holds the 11 pairs sharing one residue. \textbf{(a)} At initialization the trivial self-overlaps form the diagonal, while off-diagonal overlaps are scattered without residue-class block structure. \textbf{(b)} After crystallization the pairs form 11 tight, nearly orthogonal state clusters, one per residue class. \textbf{(c)} The two classes outlined in orange in (b), labeled by individual input pairs, showing 22 equations organized into two tight clusters.}}
    \label{fig:group_collapse}
\end{figure*}

\textcolor{black}{Using modular addition with $p=11$ as a representative case, we investigate what representation a crystallized UQT converges to. The two operands act sequentially on the same four embedding qubits, so the circuit prepares $|\psi(a,b)\rangle = E_b E_a|0000\rangle$, with $E_t$ the embedding unitary for token $t$. Figure~\ref{fig:group_collapse} shows the overlaps $|\langle\psi(a,b)|\psi(a',b')\rangle|^2$ for all 121 ordered pairs. At initialization, the mean within-class overlap of 0.057 is comparable to the between-class value of 0.062, with no residue-class block structure. After crystallization, the pairs form 11 tight, nearly orthogonal state clusters, one per residue class, with mean within-class overlap 0.99996 and no between-class state-pair overlap above 0.050. Importantly, this residue-class cluster is found by optimization rather than supplied by construction.}

\section{Why the proposed architecture crystallizes}
\label{app:theory}
We now provide a constructive argument showing that the proposed quantum attention architecture can represent modular addition exactly when its embedding operators are chosen to respect the symmetry of the task. \textcolor{black}{While the following discussion is focused on modular arithmetic as an illustrative case, a unified formal treatment spanning non-Abelian algebra and linguistic compositionality remains an open direction for future theoretical work.} Consider the function
\begin{equation}
f(n,m)=n+m \pmod p,
\end{equation}
where $n,m \in \mathbb{Z}_p$ and $p\ge 2$ is the modulus. The key observation is that modular addition is the group operation of the finite cyclic group $\mathbb{Z}_p$ \cite{gromov2023grokking}. Therefore, the natural basis for representing this task is given by the characters of $\mathbb{Z}_p$, namely
\begin{equation}
\chi_k(n)=e^{2\pi i kn/p}, \qquad k=0,\dots,p-1.
\end{equation}

\noindent These functions diagonalize the translation structure of the problem and satisfy
\begin{equation}
\chi_k(n)\chi_k(m)=\chi_k(n+m),
\end{equation}
where addition is understood modulo $p$. This multiplicative identity is the central reason the architecture can implement modular addition efficiently.

To analyze what happens at the embedding stage, let $\{|k\rangle\}_{k=0}^{p-1}$ denote a computational basis over a $p$-dimensional latent register. Define the token embedding operator by
\begin{equation}
U_{\mathrm{emb}}(n)|k\rangle=\chi_k(n)|k\rangle
=
e^{2\pi i kn/p}|k\rangle.
\end{equation}

\noindent Thus each token $n$ is encoded as a phase shift across spectral modes. Unlike unconstrained Euclidean embeddings, this representation is periodic and exactly compatible with modular arithmetic. By initializing the latent register in the uniform superposition, i.e.,
\begin{equation}
|\psi_0\rangle=\frac{1}{\sqrt p}\sum_{k=0}^{p-1}|k\rangle,
\end{equation}

\noindent and applying the embedding for token $n$, we have
\begin{equation}
|\psi(n)\rangle
=
U_{\mathrm{emb}}(n)|\psi_0\rangle
=
\frac{1}{\sqrt p}\sum_{k=0}^{p-1}e^{2\pi i kn/p}|k\rangle.
\end{equation}

At the composition step, applying two token embeddings sequentially yields
\begin{align}
U_{\mathrm{emb}}(m)U_{\mathrm{emb}}(n)|k\rangle
&=
e^{2\pi i km/p}e^{2\pi i kn/p}|k\rangle \\
&=
e^{2\pi i k(n+m)/p}|k\rangle.
\end{align}

\noindent Therefore,
\begin{equation}
U_{\mathrm{emb}}(m)U_{\mathrm{emb}}(n)|\psi_0\rangle
=
\frac{1}{\sqrt p}\sum_{k=0}^{p-1}e^{2\pi i k(n+m)/p}|k\rangle.
\end{equation}

\noindent The modular sum is thus accumulated directly in phase space. No learned nonlinear arithmetic rule is required; the group structure of the task is implemented by construction.

Finally, at the readout stage, to decode the symbolic output, we can apply the IQFT ($F_p^\dagger$):
\begin{equation}
F_p^\dagger |k\rangle
=
\frac{1}{\sqrt p}\sum_{q=0}^{p-1}e^{-2\pi i kq/p}|q\rangle.
\end{equation}

\noindent Then,
\begin{align}
&F_p^\dagger
\left(
\frac{1}{\sqrt p}\sum_{k=0}^{p-1}e^{2\pi i k(n+m)/p}|k\rangle
\right) \notag \\
&\quad=
\sum_{q=0}^{p-1}
\left[
\frac{1}{p}\sum_{k=0}^{p-1}e^{2\pi i k(n+m-q)/p}
\right]|q\rangle.
\end{align}

\noindent Using orthogonality of characters,
\begin{equation}
\sum_{k=0}^{p-1}e^{2\pi i ks/p}=p\,\delta_{s,0 \, (\mathrm{mod}\, p)},
\end{equation}
we obtain
\begin{equation}
F_p^\dagger U_{\mathrm{emb}}(m)U_{\mathrm{emb}}(n)|\psi_0\rangle
=
|n+m \pmod p\rangle.
\end{equation}

Accordingly, measurement returns the correct modular sum with probability one. While this analytical construction relies on an explicitly programmed IQFT and exact character phase shifts, it serves as a foundational existence proof: the exact geometry of modular arithmetic is natively resolvable within a multi-qubit Hilbert space. \textcolor{black}{Generic classical architectures do not encode this particular periodic construction explicitly, so training must discover a suitable representation from the data.}

Note that in our proposed UQT, we do not hardcode the IQFT nor the exact $\mathbb{Z}_p$ group characters. Instead, the architecture utilizes parameterized $SU(2)$ rotations ($R_x$ and $R_y$ gates) and unconstrained gradient optimization to organically discover a task-specific basis transformation. The phenomenon of crystallization we observe empirically suggests that the UQT learns a highly compressed representation of the phase-accumulation and decoding process described above, ultimately achieving the same deterministic generalization. \textcolor{black}{For $p=11$ modular addition, Appendix~\ref{app:group_structure} shows specifically that the learned pair-encoded states organize into tight residue-class clusters.} \textcolor{black}{In this experiment, the UQT converges to the target algebraic structure without the instability observed in the tested classical baselines.}

It is also important to note the key difference between the aforementioned discussion and the argument presented by Gromov \cite{gromov2023grokking}. In their work, they utilized real-valued cosine features as an approximate Fourier basis (unlike the proposed method that employs the full complex characters, $e^{2\pi i kn/p}$). The cosine formulation captures only the real projection of the harmonic structure, effectively discarding the imaginary sine component that carries phase orientation and completes the representation. As a result, the MLP must reconstruct modular relations indirectly through nonlinear interactions and training dynamics, whereas the proposed architecture preserves the full phase information from the outset, allowing modular composition to arise naturally through direct multiplication of complex phases rather than through delayed emergence of partial real-valued harmonics.

\section{Hardware execution}
\label{app:hardware}
Inference was evaluated on physical superconducting quantum processors accessed via the IBM Qiskit Runtime Service \cite{javadi2024quantum}: \texttt{ibm\_fez}\textcolor{black}{, \texttt{ibm\_marrakesh}, and \texttt{ibm\_kingston} (all featuring the 156-qubit Heron r2 architecture)}. All hardware executions utilized $8,192$ measurement shots per evaluation.

Because the physical fidelity of superconducting qubits drifts over time due to continuous environmental decoherence, it is necessary to record the exact baseline hardware metrics at the time of execution to appropriately contextualize the algorithmic noise-resilience of the UQT. The median two-qubit ($CZ$) gate infidelities across the processors ranged from \textcolor{black}{$1.82 \times 10^{-3}$} to $2.69 \times 10^{-3}$, and median readout assignment errors ranged from \textcolor{black}{$8.91 \times 10^{-3}$} to $1.52 \times 10^{-2}$. Median $T_1$ relaxation times were bounded between $139.63 \, \mu\text{s}$ and \textcolor{black}{$280.29 \, \mu\text{s}$}, while $T_2$ dephasing times ranged from $96.32 \, \mu\text{s}$ to \textcolor{black}{$138.43 \, \mu\text{s}$}. Because the UQT does not utilize quantum error correction, the successful, unmitigated hardware classification achieved in our results confirms that the learned geometric phase-interference is robust enough to overcome native environmental decoherence and localized gate degradation.

\bibliographystyle{apsrev4-2}
\bibliography{Bibliography}

\end{document}